\documentclass[lettersize,journal]{IEEEtran}
\usepackage{amsmath,amsfonts}
\usepackage{algorithmic}
\usepackage{array}
\usepackage[caption=false,font=footnotesize,labelfont=rm,textfont=rm, subrefformat=parens]{subfig}
\usepackage{textcomp}
\usepackage{stfloats}
\renewcommand\ref[1]{\subref*{#1}}
\usepackage{url}
\usepackage{verbatim}
\usepackage{graphicx}
\usepackage{booktabs}%提供命令\toprule、\midrule、\bottomrule
\usepackage{amssymb}
\usepackage{mathtools} 
\usepackage{amsmath}
\usepackage{multirow}
\usepackage{graphicx}
\usepackage{booktabs}
\usepackage{color}
\usepackage{hyperref}
\graphicspath{{images/}}
\graphicspath{{line_graph/}}
\graphicspath{{curve/}}
\graphicspath{{heatmap/}}
\graphicspath{{authors/}}
\usepackage[ruled,linesnumbered]{algorithm2e}
\def\BibTeX{{\rm B\kern-.05em{\sc i\kern-.025em b}\kern-.08em
    T\kern-.1667em\lower.7ex\hbox{E}\kern-.125emX}}
\usepackage{cite}
\begin{document}
\title{Single Node Injection Label Specificity Attack on Graph Neural Networks via Reinforcement Learning}
%Deep Reinforcement Learning-Based Label Specificity Attack Against Graph Neural Networks
% \author{Dayuan Chen
% \thanks{Manuscript created October, 2020; This work was developed by the IEEE Publication Technology Department. This work is distributed under the \LaTeX \ Project Public License (LPPL) ( http://www.latex-project.org/ ) version 1.3. A copy of the LPPL, version 1.3, is included in the base \LaTeX \ documentation of all distributions of \LaTeX \ released 2003/12/01 or later. The opinions expressed here are entirely that of the author. No warranty is expressed or implied. User assumes all risk.}}
% 加在开头
\author{
  Dayuan~Chen,
  Jian~Zhang,
  Yuqian~Lv,
  Jinhuan~Wang,
  Hongjie~Ni,
  Shanqing~Yu,
  Zhen~Wang,
  and~Qi~Xuan,~\IEEEmembership{Senior~Member,~IEEE}% <-this % stops a space
  %\thanks{
  %This work was supported in part by the Key R\&D Program of Zhejiang under Grant 2022C01018, by the National Natural Science Foundation of China under Grants 61973273 and U21B2001, by the National Key R\&D Program of China under Grant 2020YFB1006104, and by The Major Key Project of PCL under Grants PCL2022A03, PCL2021A02, and PCL2021A09, by the National Natural Science Foundation of China under Grants 62103374, by the the key R&D projects in Zhejiang Province under Grant 2021C01117.
 % }
    \thanks{
    D. Chen, Y. lv, J. Wang and S. Yu are with the Institute of Cyberspace Security, College of Information Engineering, Zhejiang University of Technology, Hang Zhou, 31000, China (e-mail: dayuanchen@zjut.edu.cn; jhwang@zjut.edu.cn; yushanqing@zjut.edu.cn).
  }
  \thanks{
  J. Zhang is with the School of Cyberspace, Hangzhou Dianzi University, Hangzhou 31000, China (e-mail: zhangjian-hdu@hdu.edu.cn).
  }
  \thanks{
  H. Ni is with the College of Information Engineering, Zhejiang University of Technology, Hang Zhou, 31000, China (e-mail: zdfynhj@zjut.edu.cn).
  }
  \thanks{
  Z. Wang is with the School of Cyberspace, and ZhuoYue Honors College, Hangzhou Dianzi University, Hangzhou 31000, China (e-mail: wangzhen@hdu.edu.cn).
  }% <-this % stops a space
  \thanks{
    Q. Xuan is with the Institute of Cyberspace Security, College of Information Engineering, Zhejiang University of Technology, Hangzhou, 310023, China, with the PCL Research Center of Networks and Communications, PengCheng Laboratory, Shenzhen, 518000, China (e-mail: xuanqi@zjut.edu.cn).
  }
}
% \markboth{
% %Journal of \LaTeX\ Class Files
% IEEE Transactions on Computational Social Systems,~Vol.~18, No.~9, September~2020}%
% {How to Use the IEEEtran \LaTeX \ Templates}

\maketitle
\begin{abstract}
Graph neural networks (GNNs) have achieved remarkable success in various real-world applications. However, recent studies highlight the vulnerability of GNNs to malicious perturbations. Previous adversaries primarily focus on graph modifications or node injections to existing graphs, yielding promising results but with notable limitations. Graph modification attack~(GMA) requires manipulation of the original graph, which is often impractical, while graph injection attack~(GIA) necessitates training a surrogate model in the black-box setting, leading to significant performance degradation due to divergence between the surrogate architecture and the actual victim model. Furthermore, most methods concentrate on a single attack goal and lack a generalizable adversary to develop distinct attack strategies for diverse goals, thus limiting precise control over victim model behavior in real-world scenarios. To address these issues, we present a gradient-free generalizable adversary that injects a single malicious node to manipulate the classification result of a target node in the black-box evasion setting. Specifically, we model the single node injection label specificity attack as a Markov Decision Process (MDP) and propose \textit{\underline{G}radient-free \underline{G}eneralizable \underline{S}ingle \underline{N}ode \underline{I}njection \underline{A}ttack}, namely G$^2$-SNIA, a reinforcement learning framework employing Proximal Policy Optimization (PPO). By directly querying the victim model, G$^2$-SNIA learns patterns from exploration to achieve diverse attack goals with extremely limited attack budgets. Through comprehensive experiments over three acknowledged benchmark datasets and four prominent GNNs in the most challenging and realistic scenario, we demonstrate the superior performance of our proposed G$^2$-SNIA over the existing state-of-the-art baselines. Moreover, by comparing G$^2$-SNIA with multiple white-box evasion baselines, we confirm its capacity to generate solutions comparable to those of the best adversaries.

\end{abstract}


\section{Introduction}
% 介绍当下主流的攻击方式及主要工作
% 介绍GNNs的应用广泛和GNNs在对抗攻击下的脆弱性
\IEEEPARstart{G}{raph} Neural Networks (GNNs), a subfield of deep learning methods, have garnered significant attention among scholars for their ability to model structured and relational data. The utilization of GNNs has demonstrated a remarkable impact in various applications of graph data mining, including node classification \cite{nodeclassification, gcn, gat, graphsage}, link prediction \cite{linkprediction,kipf2016variational, zhang2018link, zhang2020hyper}, community detection \cite{communitydetection,shchur2019overlapping, chen2017supervised, sun2021graph}, and graph classification \cite{xuan2021sgn, wang2021sampling, zhou2021evolve, le2021parameterized}.
%wu2019net
Despite their widespread adoption, recent studies have demonstrated the vulnerability of GNNs to adversarial attacks~\cite{sun2018adversarial, li2020deeprobust}. 
Imperceptible but intentionally-designed perturbations on graphs can effectively mislead GNNs to make incorrect predictions.

Pioneer attack methods against GNNs typically follow the setting of graph modification attack~(GMA)~\cite{nettack,rls2v,fga, mga, eda, lk, qattack, cdattack, rewatt, epoatk}, where adversaries can directly modify the relationships and features of existing nodes. However, these attack strategies have limited practical meaning as they require a high level of access authority over the graph, which is impractical under most circumstances. In addition to GMA, another emerging trend in adversarial research focuses on graph injection attack~(GIA)~\cite{nipa, afgsm, gnia, tdgia, clusterattack, adima, ugba, nicki}. GIA explores a more practical setting where adversaries inject new nodes into the original graph to propagate malicious perturbations, which has proven to be more effective than GMA due to its high flexibility~\cite{hao}. For instance, in a social network, adversaries do not have permission to alter the existing relationships between users, such as adding or removing friendships. However, adversaries could easily register a fake account, establish a relationship with the target user, and manipulate the behavior of the fake user to dominate the prediction results of GNNs on the target user. %\textcolor{blue}{Therefore, in this paper, we explore the extent to which GNNs' predictive behavior can be controlled by GIA.}
Therefore, in this paper, we focus on GIA when performing attacks against GNNs.

The key to GIA lies in generating appropriate malicious injected nodes as well as their features that can propagate perturbations along the graph structure to the nodes in the original graph. Several studies, such as \cite{nipa, gani}, have proposed methods for generating injection nodes using statistical or random sampling techniques and influencing existing nodes during the training phase. However, they have been found to have poor performance when directly applied during the inference phase. Additionally, some researches \cite{afgsm, gnia, tdgia, gb_fgsm} leverage the gradient of the victim model or a surrogate model to generate injection nodes and implement an evasion attack during the inference phase. These methods perform excellently in the white-box evasion setting, but their performance decreases in the black-box evasion setting due to diverging of surrogate architecture and the actual victim model. The phenomenon of performance decreasing is especially pronounced in graphs with discrete node features because the features of the injected node are usually designed as a binary vector to ensure its imperceptibility, resulting in limited disturbance capability. Therefore, we focus on how to select the most effective features of injection nodes within the constraints of a limited attack budget in discrete feature space. Due to the weakness of gradient-based methods, some novel gradient-free methods have emerged. Ju et al. \cite{g2a2c} creates a node generator through Advantage Actor-Critic (A2C) \cite{a2c} algorithm in the black-box evasion setting, but it focuses on global attack and cannot generate efficient vicious nodes for different tasks (i.e. different target nodes and targeted labels). Therefore, when conducting GIA in discrete feature space, adversaries must consider the followings: (1) \textbf{Effectiveness}. How to generate malicious features for the injected node to implement an attack. (2) \textbf{Efficiency}. The resulting combinatorial optimization problem is NP-hard, so how can one efficiently search for a good suboptimal solution? (3) \textbf{Generalizability}. How to design a generalizable attack algorithm for misleading the victim model into assigning specific labels to different target nodes?

Besides the above three considerations, in this work, we concentrate on the most challenging and realistic scenario, where the adversary is limited to injecting only one malicious node to control the classification result of a single target node in the black-box evasion setting, namely single node injection label specificity attack.In this scenario, the adversary only has access to the connection relationships between nodes and node features, with the permission of querying the victim model in the inference phase. we propose
 \textit{\underline{G}radient-free \underline{G}eneralizable \underline{S}ingle \underline{N}ode \underline{I}njection \underline{A}ttack}, namely G$^2$-SNIA,
%\textcolor{magenta}{G$^2$-SNIA, a gradient-free generalizable attack algorithm (e.g., a \textit{Gradient-free Generalizable Single Node Injection Attack})} that can be applied to
to handle a multitude of diverse attack goals (i.e., varying target nodes and targeted labels) in the black-box evasion setting. Our approach adopts a direct attack strategy that directly affects the target node through the injected node due to the aggregation process of GNNs.We represent the sequential addition of features of the injected node as a Markov Decision Process (MDP) and map the process of adding a feature to a set of discrete actions. To solve this NP-hard problem, we use the Proximal Policy Optimization (PPO)~\cite{ppo} algorithm which can improve the performance of the deep reinforcement learning~(DRL) agent by leveraging the reward function instead of the surrogate gradients. Our experiments show that the trained DRL agent performs effectively even in the presence of a large action space.
The key contributions of the paper are as follows:
\begin{itemize}
    \item  This study is the pioneering work that investigates graph injection attack in the black-box evasion setting for a diverse range of attack goals, without relying on surrogate gradient information.
    \item We meticulously formulate the black-box single node injection label specificity attack as an MDP. To dominate the predictions of GNNs trained on graphs with discrete node features, we propose G$^2$-SNIA, a novel gradient-free generalizable attack algorithm based on a reinforcement learning framework to generate effective yet imperceptible perturbations for various attack goals. 
    \item With comprehensive experiments over three acknowledged benchmark datasets and four renowned GNNs in the black-box evasion setting, we demonstrate G$^2$-SNIA outperforms the current state-of-the-art attack methods in terms of attack effectiveness. Specifically, we achieve an average improvement of approximately 5\% over the best baselines in terms of attack success rate.
    \item In addition, we compare the performance of G$^2$-SNIA with several baselines that operate within the white-box evasion setting. Our experimental results indicate that even in black-box conditions, G$^2$-SNIA is capable of producing solutions that are comparable to those generated by the baselines. This highlights the efficacy of G$^2$-SNIA in achieving successful attacks despite the absence of knowledge about the victim GNNs.
\end{itemize}

The remainder of the paper is organized as follows: In Section \ref{RW}, we review relevant literature on adversarial attacks on GNNs and graph injection attack on GNNs. Section \ref{Preliminaries} provides a formal definition of the single node injection label specificity attack problem. Our proposed solution, G$^2$-SNIA, is presented in Section \ref{Method}. Section \ref{experiments} describes our experimental results. Finally, Section \ref{conclusion} concludes with a summary and an outline of promising directions for future work.
%and outlines promising directions for future work.
% This document applies to version 1.8b of IEEEtran. 

% The IEEEtran template package contains the following example files: 
% \begin{list}{}{}
% \item{bare\_jrnl.tex}
% \item{bare\_conf.tex}
% \item{bare\_jrnl\_compsoc.tex}
% \item{bare\_conf\_compsoc.tex}
% \item{bare\_jrnl\_comsoc.tex}
% \end{list}
% These are ``bare bones" templates to quickly understand the document structure.  

% It is assumed that the reader has a basic working knowledge of \LaTeX. Those who are new to \LaTeX \ are encouraged to read Tobias Oetiker's ``The Not So Short Introduction to \LaTeX '', available at: \url{http://tug.ctan.org/info/lshort/english/lshort.pdf} which provides an overview of working with \LaTeX.   

\section{Related Work} \label{RW}
\subsection{Adversarial Attacks on GNNs}
 In most existing studies, adversaries are launched by modifying edges and features of the original graph \cite{nettack,fga, mga, rls2v, cdattack, epoatk, dsem}, significantly degrading the performance of GNN models. Nettack \cite{nettack} modifies node features and the graph structure guided by the surrogate gradient. RL-S2V \cite{rls2v} uses reinforcement learning to flip edges. MGA \cite{mga} (Momentum Gradient Attack) combines the momentum gradient algorithm to achieve better attack effects. DSEM \cite{dsem} assigns a specific label to a target node by adding the node with the highest degree from the targeted label node set in the entire graph to the target node in the black-box evasion setting. However, pioneer graph modification attack (GMA) methods require high authority making them infeasible to implement in real-world scenarios. 
\subsection{Graph Injection Attack on GNNs}
To address the dilemma, research on graph injection attack (GIA), a more realistic attack type, has surged. GIA involves the injection of vicious nodes, rather than modifying the original graph. NIPA \cite{nipa} generates injected node features by adding Gaussian noise to the mean of node features and executes attack during the training phase. GANI \cite{gani} computes the frequency of feature occurrences among nodes sharing the same label and selects the most frequent feature as the injected node feature to poison nodes in the original graph. GB-FGSM \cite{gb_fgsm} is gradient-based greedy method that calculate the gradient of the victim model in the white-box evasion setting and the gradient of the surrogate model in the black-box evasion setting to generate malicious injected nodes. G-NIA \cite{gnia} employs a neural network to model the attack process for preserving learned patterns and implements attacks in the inference phase. In work that is most closely related to ours, G$^2$A2C \cite{g2a2c} leverages a reinforcement learning algorithm to generate injected node features in the black-box evasion setting. However, there are several key differences between G$^2$A2C and our proposed method, G$^2$-SNIA: (1) G$^2$A2C focuses on global attack, whereas G$^2$-SNIA is a generalizable adversary that aims to mislead the victim model into assigning specific labels to different target nodes. (2) G$^2$A2C conducts attacks by injecting multiple nodes, while G$^2$-SNIA allows for the injection of only one extra node and one edge connected to the target node in order to preserve the predictions of the victim model for other nodes in the original graph. (3) There are major differences in the reward functions used by G$^2$A2C and G$^2$-SNIA. (4) G$^2$A2C employs off-policy reinforcement learning algorithm, while we adopt on-policy method, which can significantly reduce memory overhead.

\section{Preliminaries} \label{Preliminaries}
\subsection{GNNs for Node Classification Task}
\textbf{Node Classification}. Following the standard notations in the literature \cite{sun2018adversarial}, we define $G=(V,E,\mathbf{X})$ as an attributed graph with $N$ nodes, where $V=\{v_i \ | \ i=1,2,\dots,N\}$ represents the set of nodes, $E=\{e_{ij}=(v_i, v_j)\ |\ v_i, v_j \in V\}$ represents the set of edges between nodes, and $\mathbf{X}=[\mathbf{x}_1,\mathbf{x}_2,\dots,\mathbf{x}_N]^T \in \{0, 1\}^{N \times F}$ is the feature matrix, where $\mathbf{x}_i$ denotes the feature vector for node $v_i$ and $F$ is the feature dimension. The adjacency matrix $\mathbf{A} \in \{0, 1\}^{N \times N}$ contains the information about the node connections, where each component $\mathbf{A}_{ij}$ represents whether the edge $e_{ij}$ exists in the graph. For simplicity, we use $G=(\mathbf{A}, \mathbf{X})$ to refer to an unweighted and undirected attributed graph in this paper. We assign a ground truth label $y_i \in \mathcal{Y}=\{1,2,\dots,Y\}$ to node $v_i$ in the graph, where $Y$ denotes the total number of labels.
In many real-world scenarios, label information is only available for a limited number of nodes. Therefore, we partition the nodes in the graph into training and test sets. 
The training set $V_{L} \subset V$ consists of labeled nodes where each node $v_i \in V_{L}$ is associated with $y_i \in \mathcal{Y}$. 
The test set $V_{U} \subset V$ consists of unlabeled nodes whose labels need to be predicted, which ensures that $V_{L} \cap V_{U} = \emptyset$. We also define a set of target nodes $V_{tar} \subset V_U$ that each node will be attacked.
The goal of node classification is to assign labels to each unlabeled node $v \in V_{U}$ by a classifier $\mathbf{Z}^{(G)}=f_{\theta}(G)$, where $ \mathbf{Z}^{(G)} \in \mathbb{R}^{N \times Y}$ represents the probability distribution matrix, and $\theta = \{ \mathbf{W}^{(1)},\mathbf{W}^{(2)},\dots\}$ represents parameter set of classifier. 

\textbf{Graph Neural Networks}. A typical GNN layer applied to a target node $v_t$ can be expressed as an aggregation process:

\begin{equation}
    \mathbf{h}^{l+1}_{v_t} = \phi(
    \alpha^l_{v_t,v_t} \mathbf{h}^{l}_{v_t} + 
    \sum\limits_{v_u \in V \setminus v_t}
    \alpha^l_{v_t,v_u} \mathbf{h}^{l}_{v_u}),
\end{equation}
where $\phi(\cdot): \mathbb{R}^{F_{in}} \to \mathbb{R}^{F_{out}}$ is a vector-valued function and $\alpha^l_{v_i,v_j}$ represents the weight assigned to the feature $\mathbf{h}^l_{v_j}$ of node $v_j$ during the aggregation process on node $v_i$. We have $\mathbf{h}^0_{v_i}=\mathbf{x}_i$ as initial.
Taking the GCN as an example: 
\begin{equation}
    \mathbf{h}^{l+1}_{v_t} = \phi(
    \frac{1}{\Tilde{d}_{{v_t}}} \mathbf{h}^{l}_{v_t} + 
    \sum\limits_{v_u \in \mathcal{N}_1(v_t)}
    \frac{1}{\sqrt{\tilde{d}_{v_t} \tilde{d}_{v_u}}} \mathbf{h}^{l}_{v_u}), 
\end{equation}
where $\tilde{d}_{v_i} = \tilde{\mathbf{D}}_{ii}$, $\tilde{\mathbf{D}} \in \mathbb{R}^{N \times N}$ is the degree matrix of $\tilde{\mathbf{A}} \in \mathbb{R}^{N \times N}$, $\tilde{\mathbf{A}} = \mathbf{A} + \mathbf{I}$ is the adjacency matrix with self-loops,  $\mathcal{N}_1(v_t)$ is the one-hop neighbors of node $v_t$. Here,  $\phi(\mathbf{h})=\sigma(\mathbf{hW})$, where $\mathbf{h} \in \mathbb{R}^{F_{in}}$ is an input vector, $\sigma(\cdot)$ is the activation function and $\mathbf{W} \in \mathbb{R}^{F_{in} \times F_{out}}$ is a weight matrix for transformation.

\begin{table}[!t]
\caption{Notations and Explanations}
\label{Notations}
\begin{tabular*}{\hsize}{@{}@{\extracolsep{\fill}}ll@{}}
\toprule[0.5mm]
\textbf{Notation} & \textbf{Explanation} \\
\toprule[0.25mm]
    $V_L$             &         Labeled node set            \\
    $V_U$            &          Unlabeled node set, $V \setminus V_L$            \\
    $V_{tar}$         &         Target node set, $V_{tar} \setminus V_U$ \\
    $\mathcal{Y}$       &        Label set \\
    $\hat{G}$             &      Adversarial graph \\
    $\mathbf{Z}^{(G)}$  &    Probability distribution of original graph $G$ \\
    $\hat{\mathbf{Z}}^{(\hat{G})}$  &    Probability distribution of adversarial graph $\hat{G}$ \\
    $v_t$               &       Target node in $V_{tar}$\\
    $y_t$               &       Targeted label in $\mathcal{Y}$ \\
    $\Delta$            &       Attack budget \\
\toprule[0.25mm]
    $\hat{G}_t$             &      Adversarial graph at time step $t$ \\
    $(c_{v_t}, \hat{G}_t)$  &      Target node's subgraph of its associated graph  $\hat{G}_t$ \\
    $\mathbf{n}_{v_t}$                        &      Node representation of target node $v_t$    \\ 
    $\mathbf{l}_{y_t}$      &      Label representation of targeted label $y_t$ \\ 
    $\mathbf{\mathcal{E}}_t$ &     State embedding at time step $t$ \\
    $s_t$                   &      State at time step $t$ \\
    $a_t$                   &      Action at time step $t$ \\
    $\mathbf{m}_t$          &      Mask at time step $t$ \\
    $r_{t}$               &      Reward function at time step $t$ \\
    $\pi_{\theta}(\cdot|s_t)$        &      Policy of state to action distribution at time step $t$ \\
    $V_{\theta}(s_t)$                &      Value function at time step $t$ \\ 
\toprule[0.25mm]
\end{tabular*}
\end{table}
\subsection{Problem Definition}
The goal of the single node injection label specificity attack is to control the classification result of a single target node in extremely limited setting, while minimizing the impact on other nodes in the graph. We define the adversarial graph $\hat{G}=(\hat{\mathbf{A}}, \hat{\mathbf{X}})$ as the original graph $G$ after undergoing small perturbations. The new graph can be expressed as:
\begin{equation}
    \hat{\mathbf{A}} = 
    \begin{bmatrix} 
    \mathbf{A} & \hat{\mathbf{e}} \\
    \hat{\mathbf{e}}^T & 0
    \end{bmatrix},
     \hat{\mathbf{X}} = 
    \begin{bmatrix} 
    \mathbf{X} \\
    \hat{\mathbf{x}}^T
    \end{bmatrix}.
\end{equation}
Here $\hat{\mathbf{e}} \in \{0, 1\}^N$ represents the relationship between original nodes and the injected node where $\hat{\mathbf{e}}_i=1$ if the injected node $v_{inj}$ is connected to the original node $v_i \in V$. Moreover, $\hat{\mathbf{x}} \in \{0, 1\}^N$ denotes the feature vector for the injected node $v_{inj}$. We formalize the objective function as:
\begin{align}
\begin{aligned}
     %    \mathop{\min}\limits_{\hat{G}}  \quad& \mathcal{L}_{atk}(v_t, y_t, \hat{G}) = -\ln \hat{\mathbf{Z}}^{(\hat{G})}_{vt,yt} \\
     % s.t. \quad & \left|\hat{\mathbf{e}}\right|_0 \leq \Delta_{e}, \left|\hat{\mathbf{x}}\right|_0 \leq \Delta_{f} \\
     % \theta^* =& \mathop{\arg\min}\limits_{\theta} \sum\limits_{v_i \in V_L} -\ln \mathbf{Z}^{(G)}_{v_i, y_i}
     \mathop{\min}\limits_{\hat{G}} \qquad &\mathcal{L}_{atk}(v_t, y_t, \hat{G}) = -\ln \hat{\mathbf{Z}}^{(\hat{G})}_{vt,yt} \\
     s.t. \qquad &\theta^* = \mathop{\arg\min}\limits_{\theta} \sum\limits_{v_i \in V_L} -\ln \mathbf{Z}^{(G)}_{v_i, y_i} \\
                 &\lVert\hat{\mathbf{e}}\rVert_0 \leq \Delta_{e} \\ &\lVert\hat{\mathbf{x}}\rVert_0 \leq \Delta_{f},
\end{aligned}
\end{align}
where $v_t \in V$ is the target node to be attacked, $y_t$ is the targeted label, $\hat{G}$ is the perturbed graph, $\hat{\mathbf{Z}}^{(\hat{G})}=f_{\theta*}(\hat{G})$ is the probability distribution matrix after perturbed, $\lVert\cdot\rVert_0$ is the $L_0$ norm, $\theta^*$ is the parameter set of classifier with optimized on the original graph, and $\mathcal{L}_{atk}$ is the negative cross-entropy loss with respect to the targeted label $y_t$ for the target node $v_t$. This means that the objective of adversary is to maximize the confidence level of targeted label on the target node. In order to achieve attack goal, the injected node only allow to connect to the target node i.e., $\hat{\mathbf{e}}_i = 1$ if $v_i = v_t$ and $\Delta_e = 1$. Thus, the perturbation is limited to the generation of the malicious feature of the injected node within the feature budget $\Delta_f$.

For the convenience of the reader, the notations used in the paper are summarized in Table \ref{Notations}.
\section{Proposed Method} \label{Method}
%The node injection attack leverages the aggregation process to propagate malicious features onto the target node without requiring retraining. Hence, generating an appropriate feature vector within an extremely large discrete space $O(C^{F}_{\Delta_F})$ is critical, but this combinatorial optimization problem cannot be exhaustively enumerated. Deep reinforcement learning offers a promising approach for solving this combinatorial optimization problem.

%We design and propose \textit{\underline{G}radient-free \underline{G}eneralizable \underline{S}ingle \underline{N}ode \underline{I}njection \underline{A}ttack}, namely G$^2$-SNIA, to perform the single node injection label specificity attack in black-box evasion setting, which brings the generalization ability while keeping the high attack performance. Fig. \ref{framework} illustrates the overall framework of the proposed model G$^2$-SNIA. The key idea behind our proposed framework is to use a DRL agent to iteratively perform actions aimed at fooling the victim model. More specifically, Given a graph $G$, a target node $v_t$, a targeted label $y_t$ and an initial injected node $v_{inj}$  connected to the target node with $\hat{\mathbf{x}} \in \{0\}^F$ , the DRL agent adds a feature (i.e., a word from the bag-of-words) step-by-step to the injected node. We now proceed to describe the DRL environment and  PPO algorithm to accomplish attack goal.

\begin{figure*}[!t]
    \centering
    \includegraphics[width=\textwidth]{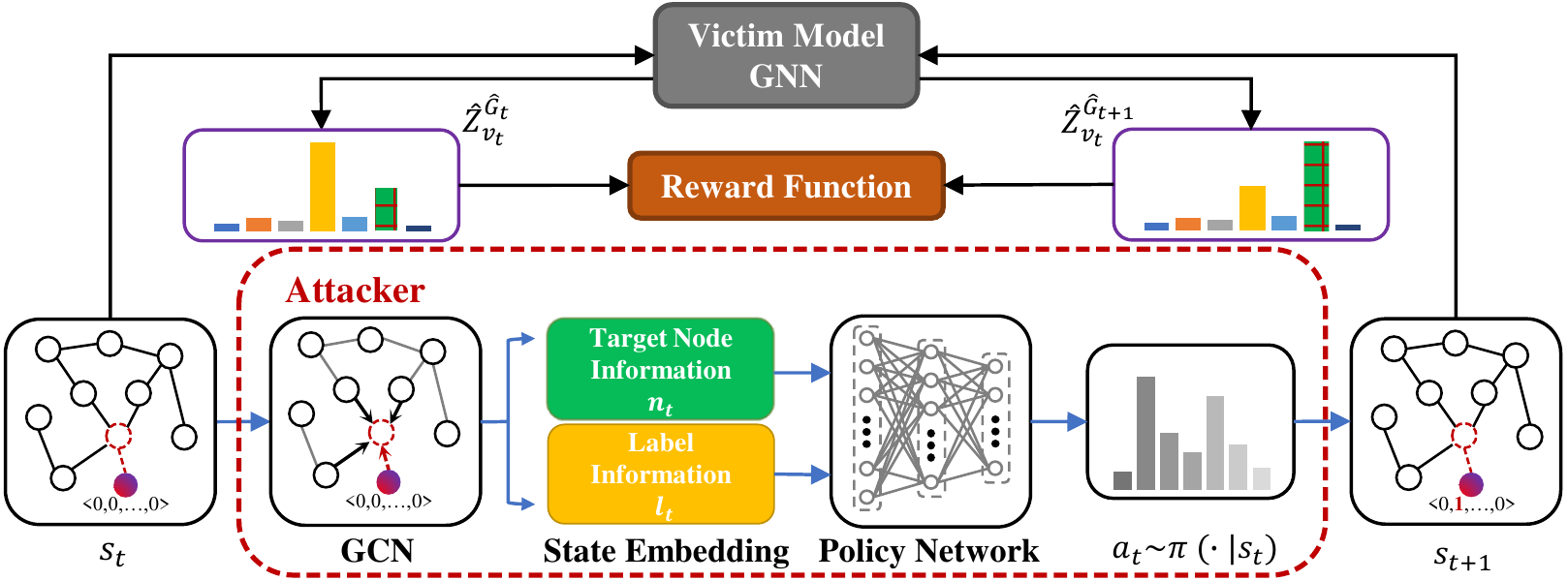}
    \label{framework}
    \caption{The overall framework of G$^2$-SNIA.}
\end{figure*}

We present the \textit{\underline{G}radient-free \underline{G}eneralizable \underline{S}ingle \underline{N}ode \underline{I}njection \underline{A}ttack} (G$^2$-SNIA), a novel approach for performing single node injection label specificity attack in the black-box evasion setting. Our proposed method combines generalization capabilities with high attack performance. The overall framework of G$^2$-SNIA is illustrated in Fig.~\ref{framework}. The key idea behind our proposed framework is to use a DRL agent to iteratively perform actions to fool the victim model. More specifically, given a graph $G$, a target node $v_t$ with targeted label $y_t$, and an initial injected node $v_{inj}$ connected to $v_t$ with initial feature vector $\hat{\mathbf{x}} \in \{0\}^F$, the DRL agent adds a feature (i.e., a word from the bag-of-words) to the injected node in a step-by-step manner. In the following sections, we describe the DRL environment and the PPO algorithm used to train the DRL agent.

\subsection{Environment}
We model the proposed single node injection label specificity attack using a model-free Markov Decision Process (MDP) $\left< \mathcal{S},\mathcal{A}, \mathcal{P}, \mathcal{R},\gamma \right>$. Here, $\mathcal{S}$ refers to the set of states, $\mathcal{A}$ denotes the set of actions, $\mathcal{R}$ is the reward function, and $\gamma < 1$ is the discount factor that represents the relative importance of immediate rewards versus future rewards. In model-free methods, the state transition probability function $\mathcal{P}$ is often considered unknown, and the objective of DRL agent is to learn the optimal policy directly without necessarily learning the complete model of the environment.

\textbf{State}. The state $s_t$ contains the intermediate adversarial graph $\hat{G}_t$ at time step $t$, a target node $v_t$, and a targeted label $y_t$. To capture information about the target node $v_t$ in the non-Euclidean structure of the adversarial graph $\hat{G}_t$, we extract the 2-hop subgraph $(c_{v_t}, \hat{G}_t)$ of the target node $v_t$, which combines both topological and feature information to represent node information. Then we group all nodes in original graph based on their classification results by the classifier and pooling the node information in the each group as the targeted label information $\mathbf{l}_{y_t}$.

\textbf{Action}. In our attack scenario, we constrain the injected node $v_{inj}$ to connect exclusively with the target node $v_t$, thereby enabling us to focus on generating the feature vector for the injected node $v_t$. To this end, we employ an iterative strategy whereby the action taken at time step $t$, denoted by $a_t$, involves adding a feature to the injected node $v_t$ (i.e., $\hat{\mathbf{x}}_i = 1$ if the DRL agent selects a feature $i$). In addition, by employing invalid action masking technique, we define a mask $\mathbf{m}_t$ at time step $t$ to prevent repeated selection of actions within an episode. The trajectory of our proposed model-free MDP is given by $(s_0,a_0,r_0,s_1,a_1,\dots,s_{T-1},a_{T-1},r_{T-1}, s_T)$, where $s_T$ denotes the terminal state and $r_{t}$ represents the intermediate reward that depends on the current state $s_t$ and the decision $a_t$ made at time step $t$.

\textbf{Reward}. A well-designed reward function is essential to ensuring the success of a deep reinforcement learning algorithm in achieving convergence. In light of the prolonged learning trajectory for the DRL agent in its environment, providing diverse intermediate rewards to the DRL agent at various states along the trajectory can be more advantageous than furnishing a sparse reward on the terminal state or endowing multiple inefficient rewards such as $\pm 1$ throughout the trajectory. This is due to the fact that diverse intermediate rewards can provide invaluable guidance to the DRL agent as it explores the environment, leading to more efficient decisions and ultimately improving its overall performance. The proposed guiding reward denoted as $r_{t}$ at time step $t$ is defined as follows:
\begin{align}
\label{reward}
\begin{aligned}
    r_{t} &= \mathcal{L}_{atk}(v_t,y_t,\hat{G}_{t}) - \mathcal{L}_{atk}(v_t,y_t,\hat{G}_{t+1}) \\
            &= -\ln \hat{\mathbf{Z}}^{(\hat{G}_{t})}_{v_t, y_t} + \ln \hat{\mathbf{Z}}^{(\hat{G}_{t+1})}_{v_t, y_t}.
\end{aligned}
\end{align}
Here, we leverage the discrepancy in entropy as the reward function, which is computed by the classifier on the adversarial graph across consecutive time intervals. This design guides the DRL agent towards taking actions that maximize the reduction of entropy in each step along its trajectory.

\textbf{Terminal}. The attack budget $\Delta_f$ is imposed as a constraint on the number of allowed malicious features added to ensure the imperceptibility of the injected node $v_{inj}$. Thus, when the DRL agent has added the maximum number of features ($\Delta_f$) to the injected node $v_{inj}$, it ceases to take any further actions. At the terminal state $s_T$, the adversarial graph $\hat{G}_{T}$ comprises only an additional injected node and an extra edge connecting the injected node to the target node, in addition to those already present in the original graph $G$.

\subsection{Single Node Injection Label Specificity Attack via PPO}

% 未优化部分
\textbf{Embedding of State}. In the aforementioned context, the state $s_t$ encompasses the intermediate adversarial graph $\hat{G}_t$ at time step $t$, along with the target node $v_t$ and the targeted label $y_t$. To extract the target node information, G$^2$-SNIA utilizes graph convolutional aggregation on the 2-hop subgraph $(c_t, \hat{G}t)$ of the target node $v_t$, which integrates both the topological and feature information. The representation of the target node $v_t$ denoted as $\mathbf{n}_{v_t} \in \mathbb{R}^F$ is defined as follows: 
\begin{equation}
\label{node representation}
    \mathbf{n}_{v_t} = \frac{1}{\hat{d}_{{v_t}}} \mathbf{X}_{v_t} +  
    \sum\limits_{v_u \in \hat{\mathcal{N}}_1(v_t) } \frac{1}{\sqrt{\hat{d}_{v_t} \hat{d}_{v_u}}} \mathbf{X}_{v_u},
     %\frac{1}{\sqrt{\tilde{d}_{v_t}}} \hat{\mathbf{x}}
\end{equation}
where $\hat{d}_i$ is the degree of node $i$ in subgraph $(c_t, \hat{G}_t)$ with self-loop, and $\hat{\mathcal{N}}_1(v_t)$ refers to the set of one-hop neighbors of the target node $v_t$ in the same 2-hop subgraph.

To improve the representation of the targeted label $y_t$ and ensure dimension unification of label representation and node representation, a strategy of utilizing multiple node representations to represent labels is employed instead of using one-hot encoding. This involves grouping all nodes in the original graph $G$ based on their classification results obtained from the classifier, and pooling the node information within each group. The representation of the targeted label $y_t$ denoted as $\mathbf{l}_{y_t} \in \mathbb{R}^F$ is defined as follows:
\begin{equation}
\label{label representation}
     \mathbf{l}_{y_t} = \mathrm{Pool}(\{\mathbf{H}_{v_i}\ | \ {v_i \in L_{y_t}}\}).
\end{equation}
Here, $L_{y_t} = \{ v_i \  | \ v_i \in V,\mathop{\arg\max}\limits_{y \in \mathcal{Y}} \mathbf{Z}^{(G)}_{v_i, y} = y_t\}$  denotes the set of nodes whose label is assigned to $y_t$ by the classifier on the original graph $G$. The matrix $\mathbf{H}=\tilde{\mathbf{D}}^{-\frac{1}{2}} \tilde{\mathbf{A}} \tilde{\mathbf{D}}^{-\frac{1}{2}}\mathbf{X} \in \mathbb{R}^{N \times F}$ represents the nodes in the original graph $G$ using graph convolutional aggregation. $Pool$ is a pooling function and we adopt mean pooling in this work.

In summary, the embedding of state at time step $t$ is represented by $\mathbf{\mathcal{E}}_t \in \mathbb{R}^{2F}$, which can be expressed as follows:
\begin{equation}
\label{state representation}
    \mathbf{\mathcal{E}}_t = \mathrm{Concat}(\mathbf{n}_{v_t}, \mathbf{l}_{y_t}),
\end{equation}
where $\mathrm{Concat}$ denotes the concatenation operation of vectors, $\mathbf{n}_{v_t}$ is the representation of the target node $v_t$ and $\mathbf{l}_{y_t}$ is the representation of the targeted label $y_t$.

\textbf{Policy Network}. We leverage a multilayer perceptron (MLP) as the policy network to generate an unbounded vector. It is not appropriate to utilize this vector directly as a probability distribution of actions even after applying the $Softmax$ function due to the potential for repeated action selection within an episode. To address this issue, the output of the policy network undergoes additional processing to ensure that it represents a valid probability distribution. Consequently, the resulting policy denoted as  $\pi_{\theta}(\cdot|s_t) \in \mathbb{R}^F$ is defined as follows:
\begin{equation}
\label{policy}
    \pi_{\theta}(\cdot|s_t) = \mathrm{Softmax}(\mathrm{MLP}(\mathbf{\mathcal{E}}_t) \odot \mathbf{m}_t+ \mathbf{\mathcal{I}} \odot 
    (1-\mathbf{m}_t)),
\end{equation}
where $\theta$ represents the parameter set of the policy network, $\mathbf{m}_t$ is the mask at time step $t$ which prevents the selection of actions that have already been taken in the current episode, $\odot$ represents Hadamard product, and $\mathbf{\mathcal{I}} \in \{-\infty\}^F$ is a constant vector with each element set to negative infinity.The $Softmax$ function is applied to the output of the policy network after processing, resulting in the probability distribution of unmasked actions $a$ given state $s_t$, denoted as $\pi_{\theta}(a|s_t)$, with the probabilities of masked actions assigned to zero to prevent repeated selection. Intriguingly, the modes of action acquisition for the training and inference phases differ starkly. Throughout the training phase, we leverage a Gumbel-Max trick \cite{gumbel} to procure an action from the probability distribution. In contrast, the inference phase employs a more straightforward approach, selecting the action with the highest probability from the given distribution.

\textbf{Value Network}. Utilizing another MLP as a value network, the value function estimates the expected return of a given state, which is employed to evaluate the desirability of actions in the present state to enable the DRL agent in improving its strategy. The value function denoted as $V_{\theta}(s_t) \in \mathbb{R}$ can be formalized as:
\begin{equation}
    V_{\theta}(s_t) = \mathrm{MLP}(\mathbf{\mathcal{E}}_t),
\end{equation}
where $\theta$ represents the parameter set of the value network.

\textbf{Training Algorithm}. The optimization of the G$^2$-SNIA model requires an iteration procedure that iteratively alternates between experience collection and parameter update until optimal performance is achieved. In the experience collection phase, the interactions between the DRL agent and the environment are recorded as $(s_t, r_t, a_t, \pi(a_t|s_t))$ in a replay buffer $\mathcal{M}$. To reduce memory and computational demands, we use the state embedding $\mathbb{\mathcal{E}}$ instead of the state, including the adversarial graph, the target node, and the targeted label. During the parameter update phase, we leverage the generalized advantage estimation (GAE) technique \cite{gae} to calculate the advantage function from the collected experiences. For example, the advantage function $A^{\pi{\theta}}_t$ for a given trajectory segment $[t_a,t_b]$ at time step index $t$ is defined as follows:
\begin{equation}
\label{advantage}
    A^{\pi_{\theta}}_t = \sum\limits_{l=0}^{t_b - t} (\gamma \lambda)^l \delta_{t+l},
\end{equation}
where $\delta_{t} = r_t + \gamma V_{\theta}(s_{t+1}) -V_{\theta}(s_{t})$ is Temporal-Difference (TD) error, and $\lambda$ is a hyper-parameter. Thus, the estimated return $G_t$ can be expressed as:
\begin{equation}
\label{return}
    G_t = A^{\pi_{\theta}}_t + V_{\theta}(s_t).
\end{equation}
Moreover, the parameter update phase of the G$^2$-SNIA model involves three key losses: policy loss $\mathcal{L}_p$, entropy loss $\mathcal{L}_e$ and value loss $\mathcal{L}_v$. To optimize the parameters of the policy $\pi_{\theta}$, we start by computing the probability ratio $r_t(\theta)$ defined as the ratio of the current policy $\pi_{\theta}(a_t|s_t)$ to the policy used during the experience collection phase $\pi_{\theta_{old}}(a_t|s_t)$, i.e., $r_t(\theta) = \frac{\pi_{\theta}(a_t|s_t)}{\pi_{\theta_{old}}(a_t|s_t)}$. Specifically, $r_t(\theta_{old}) = 1$ when the first optimization in each parameter update phase. To perform conservative policy iteration,  we define the clipped probability ratio $\hat{r}_t(\theta)$ as follows:
\begin{equation}
    \hat{r}_t(\theta) = Clip(r_t(\theta), 1-\epsilon,1+\epsilon),
\end{equation}
where $\epsilon>0$ is the clipped coefficient that controls the clip range, and $Clip(\cdot)$ is the clip function to limit the range of input, expressed as:
\begin{equation}
    Clip(x, min,max) =\left\{
	\begin{aligned}
	& min,\quad x<min\\
	& x, \quad min<x<max\\
	& max, \quad x>max\\
	\end{aligned}.
	\right
	.
 \end{equation}
 Therefore, given a tuple $(s_t, r_t, a_t, \pi(a_t|s_t)) \in \mathcal{M}$, the policy loss  $\mathcal{L}_p$ can be expressed as:
\begin{equation}
    \mathcal{L}_p= 
    -\min ( r_t(\theta)A^{\pi_{\theta_{old}}}_t,\hat{r}_t(\theta)A^{\pi_{\theta_{old}}}_t),
\end{equation}
where $A^{\pi_{\theta_{old}}}_t$ is the advantage function is computed using the policy $\pi_{\theta_{old}}$ and the value function $V_{\theta_{old}}$ during the experience collection phase. The policy loss $\mathcal{L}_p$ is designed to enhance the action selection behavior of the DRL agent by increasing the probabilities assigned to better actions. In contrast, the entropy loss $\mathcal{L}_e$ promotes exploration (i.e., the degree of unpredictability in the agent’s action selection) and is defined as follows:
\begin{equation}
    \mathcal{L}_e = \sum\limits_{a \in \mathcal{A}} \pi_{\theta}(a|s_t) \ln \pi_{\theta}(a|s_t),
\end{equation}
where $\mathcal{A}$ is the action space. Lastly, the value loss $\mathcal{L}_v$  is aimed at minimizing the discrepancy between the estimated and actual return that enables the value function to update its predictions regarding future rewards effectively and is calculated as:
\begin{equation}
    \mathcal{L}_v =\left\{
	\begin{aligned}
	& \frac{1}{2}(V_{\theta}(s_t) - G_t)^2,\quad |V_{\theta}(s_t) - G_t|<1\\
	& |V_{\theta}(s_t) - G_t|, \quad otherwise
	\end{aligned}.
	\right
	.
 \end{equation}
Therefore, given a minibatch set $\mathcal{B} \subset \mathcal{M}$, the final loss for G$^2$-SNIA is formulated as:
\begin{equation}
\label{loss}
    \mathcal{L} = -\frac{1}{|\mathcal{B}|} \sum\limits_{\mathcal{B}} \mathcal{L}_p - \mathcal{L}_v + \beta \mathcal{L}_e,
\end{equation}
where $\beta$ is the entropy coefficient that controls the exploration performance of the DRL agent. The overall training framework is summarized by Algorithm \ref{alg:algorithm1}.

\begin{algorithm}[t]
\caption{the training algorithm of framework G$^2$-SNIA}
\label{alg:algorithm1}
\KwIn{clean graph $G=(A,X)$, target node set $V_{tar}$, label set $\mathcal{Y}$, target node $v_t \in V_{tar}$, targeted label $y_t \in \mathcal{Y}$, victim model $f_{\theta^*}(\cdot)$, attack budget $\Delta_f$, training iteration $K$, experience collection steps $S$, parameter update steps $P$}
\KwOut{the general attack agent $\pi_{\theta}$}  
\BlankLine
Initialize the parameters $\theta_{old}$ of policy $\pi_{\theta_{old}}$ and value function $V_{\theta_{old}}$;
Initialize the replay  buffer $\mathcal{M}$; 
Initialize the state $s$ with target node $v_t$ and targeted label $y_t$ by random sampling; Initialize $\hat{G}$;

\While{$epoch < K$}{
    Empty replay buffer $\mathcal{M}$;
    
    \While{$step < S$}{
    Compute state embedding according to Eq.(\ref{node representation}), Eq.(\ref{label representation}) and Eq.(\ref{state representation});
    
    Compute probability distribtion of acitons according to Eq.(\ref{policy});

    Sample $a$ from $\pi_{\theta_{old}}(\cdot|s)$ by Gumbel-Max trick;

    Compute reward $r$ according to Eq.(\ref{reward});
    
    Store $(s, r, a, \pi_{\theta_{old}}(a|s))$ in $\mathcal{M}$;

    \If{ $s_{next}$ is terminal state}{
    Initialize $s_{next}$ with new target node and targeted label by random sampling;
    }
    $s \gets s_{next}$
    }
    Compute advantage function and the estimated return according to  Eq.(\ref{advantage}) and Eq.(\ref{return});

    \While{$ update\textbf{ }times < P$}{
    Sample minibatch $\mathcal{B}$ randomly from $\mathcal{M}$;

    Compute loss function according to Eq.(\ref{loss}) and update parameter $\theta$;

    $\theta_{old} \gets \theta$;
    }
}
return $\pi_{\theta_{old}}$; 
\end{algorithm}

\section{Experiments} \label{experiments}
In this section, we proceed to experiments that compare G$^2$-SNIA with several baseline methods for label specificity attacks in different attack setting. Our experiments are designed to answer the following research questions:
\begin{itemize}
% \item (\textbf{RQ1}) Given well-trained GNNs, can our proposed G$^2$-SNIA effectively change the classification result of target node to targeted label in black-box evasion setting compared with baseline methods?
\item \textbf{(RQ1)} Compared with several baselines, can G$^2$-SNIA effectively perform label specificity attack against well-trained GNNs? 
 % \item (\textbf{RQ2}) Is the result obtained from our proposed method for this NP-Hard problem a good approximate solution, and what is the efficiency of its search for this solution?
\item (\textbf{RQ2}) Can G$^2$-SNIA efficiently generate suboptimal malicious node features?
\item (\textbf{RQ3}) How does G$^2$-SNIA perform in terms of attack effectiveness under different budgets without retraining?
\end{itemize}
\subsection{Experimental Settings}
\textbf{Dataset}. 
%We conduct our experiments on three well-known public datasets: Cora \cite{cora}, Citeseer\cite{citeseer}, and DBLP\cite{dblp}. This tree datasets are citation networks where nodes are documents, edge are citation links and features are selected as the words in the document after filtering out the stop words which belong to discrete features. The detailed statistics of these datasets are shown in Table \ref{dataset}. Following the same setting in \cite{gb_fgsm} \cite{afgsm}, we only consider the largest connected component for convenience. In the experiments, we randomly split the datasets into training set (10\%), validation set (10\%) and test set (80\%). We also randomly select 1000 nodes from the testing split as a target node set denoted as $V_{tar}$ which will be employed to evaluate attack performance. In order to avoid the different performance of the surrogate model due to the different split of the dataset in the black-box attack scenario, the train/validation/test set is \textcolor{magenta}{consisitent} in training victim GNNs and the surrogate model which means the baseline methods using the surrogate model have the complete information about datasets different from usual black-box setting. 
In this work, we perform experiments on three well-known public datasets: Cora~\cite{cora}, Citeseer~\cite{citeseer}, and DBLP~\cite{dblp}. These three datasets are citation networks where nodes correspond to documents and edges represent citation links. Discrete node features are one-hot vectors which represent selected key words in corresponding documents. A comprehensive summary of these datasets is provided in Table~\ref{dataset}.
Following the same experimental setup as in~\cite{afgsm, gnia, gb_fgsm}, we focus only on the largest connected component for convenience. The datasets are randomly partitioned into training (10\%), validation (10\%), and test (80\%) sets. Additionally, we randomly select 1000 nodes from the testing set to create a target node set denoted as $V_{tar}$ to evaluate the attack performance.
To ensure consistency and eliminate performance variations due to different dataset splits in the black-box setting, the training, validation, and testing sets remain consistent while training both victim and surrogate models. It is important to note that this implies the baseline methods, which employ the surrogate model, have complete information about the datasets, which deviates from the typical black-box setting.
\begin{table}[!ht]
\caption{Statistics of the evaluation datasets}
\label{dataset}
\setlength{\tabcolsep}{1mm}
\begin{tabular*}{\hsize}{@{}@{\extracolsep{\fill}}l|c|c|c|c|l@{}}
\toprule
Dataset  & $|V|$ & $|E|$ & $F$  & $|\mathcal{Y}|$ & Frequency of classes              \\ 
\midrule
Cora     & 2485  & 5069  & 1433 & 7               & 285, 406, 726, 379, 214, 131, 344 \\
Citeseer & 2110  & 3668  & 3703 & 6               & 115, 463, 388, 304, 532, 308      \\
DBLP     & 16191 & 51913 & 1639 & 4               & 7761, 4695, 1661, 2074 \\
\bottomrule
\end{tabular*}
\end{table}

\textbf{Victim GNNs}. For all datasets, we implement four well-known GNNs as vimtim models which includes GCN \cite{gcn} and its Variants SGCN \cite{sgcn}, TAGCN \cite{tagcn} and GCNII \cite{gcnii}. It is noteworthy that the surrogate model proposed in Nettack \cite{nettack} is typically used to generate perturbations in the gray-box setting, so we chose the same model in \cite{nettack, afgsm} as the surrogate model. The accuracy of all models on test sets are shown in Table
\ref{accuracy}.
\begin{table}[!ht]
\caption{Original Classification Accuracy of GNNs}
\label{accuracy}
\setlength{\tabcolsep}{1mm}
\begin{tabular*}{\hsize}{@{}@{\extracolsep{\fill}}lccccc@{}}
\toprule
Dataset  & Surrogate & GCN   & SGCN  & TAGCN & GCNII \\ \midrule
Cora     & 83.56     & 83.71 & 84.26 & 84.62 & 84.31 \\
Citeseer & 72.41     & 73.24 & 74.25 & 73.89 & 74.04 \\
DBLP     & 83.59     & 84.00 & 83.93 & 84.14 & 84.24 \\ \bottomrule
\end{tabular*}
\end{table}

\textbf{Baseline Methods}. As single node injection attack represents an emerging type of attack, first proposed in \cite{afgsm}, only a few studies have focused on this topic, such as AFGSM \cite{afgsm}, G-NIA \cite{gnia}, and GB-FGSM \cite{gb_fgsm}. To demonstrate the effectiveness of our proposed method G$^2$-SNIA, we design a random method and a greedy-based method called MostAttr as our baselines. We also compare G$^2$-SNIA with the two state-of-the-art single node injection attack methods.
\begin{itemize}
    \item Random. We randomly sample a node from the set of nodes labeled as targeted label by the classifier on the original graph and take its feature vector as the features of the injected node to leverage aggregation process to attack the target node. 
    \item MostAttr. From the perspective of the node features, we employ a greedy-based method to generate feature vector of the malicious node. Given a targeted label, we first calculate the total non-zero appearance of features in each index among $F$ dimensions of all nodes whose classification results determined by the victim model on the original graph belong to the targeted label. Subsequently, we choose the top attack budget $\Delta_f$ feature indices with the highest appearance as the the non-zero indices of the injected node.    
    \item AFGSM~\cite{afgsm}. AFGSM is a targeted node injection poison attack that calculates the approximation of the optimal closed-form solutions to generate malicious features for the injected nodes. It sets the discrete features with the largest attack budget $\Delta_f$ approximate gradients to 1. In our attack scenario, we do not impose the constraints of co-occurrence pairs of features and improve the loss function to $-\hat{\mathbf{Z}}_{v_t, y_t}$ to maximize its attack ability as much as possible in the black-box evasion setting. It is worth noting that AFGSM has already outperformed existing attack methods (such as Nettack and Metattack) and exhibits performance very close to G-NIA in datasets with discrete features. Therefore, we only compare our proposed G$^2$-SNIA with the state-of-the-art AFGSM.
    \item GB-FGSM~\cite{gb_fgsm}. Inspired by the Fast Gradient Sign Method (FGSM) in computer vision, GB-FGSM is a single node injection backdoor attack that uses a greedy step-by-step rather than one-step optimization in the graph domain to modify the classification result of the target node to the targeted label without retraining. Each step, it only selects the most promising feature based on the maximized gradient among all features, and the iteration will continue until the budget $\Delta_f$ is exhausted.
\end{itemize}

\textbf{Parameter Settings}. 
To ensure a low attack cost, we strictly limit the attack budget $\Delta_f$ to the maximum $L_0$ norm of the feature vectors of the original nodes. We set the hyper-parameters in G$^2$-SNIA as follows: the policy network is a 6-layer MLP and the value network is a 4-layer MLP, both with 512 hidden dimensions and the $Tanh$ activation function. The discount factor $\gamma$ is set to 0.99, the clip coefficient $\epsilon$ to 0.1, the hyper-parameter $\lambda$ to 0.95, the entropy coefficient $\beta$ to 0.02, and the target steps $S$ equal to the exploration steps of the single DRL agent in an environment multiplied by the number of parallel environments. The batch size $|\mathcal{B}|$ is set to 512. We utilize the Adam optimizer with a learning rate of $2 \times 10^{-4}$ and linear decay. Additionally, we adopt early stopping with a patience of 20 evaluation epochs, with an evaluation epoch conducted after each 400 training epochs. All experiments are run on a server equipped with an Intel Xeon Gold 5218R CPU, Tesla A100, and 384GB RAM, running Linux CentOS 7.1.
\begin{table*}[!ht]
\caption{The attack success rate of single node injection label specificity attack}
\label{black}
\begin{tabular*}{\hsize}{@{}@{\extracolsep{\fill}}ccccccccccccc@{}}
%\begin{tabular}{@{}ccccccccccccc@{}}
\toprule
\multirow{2}{*}{Cora} & \multicolumn{6}{c}{GCN} & \multicolumn{6}{c}{SGCN} \\ \cmidrule(l){2-13} 
 & Clean & Random & MostAttr & AFGSM & GB-FGSM & G$^2$-SNIA & Clean & Random & MostAttr & AFGSM & GB-FGSM & G$^2$-SNIA \\ \midrule
\multicolumn{1}{c|}{$y_t$=0} & 9.3 & 13.6 & 37.5 & 69.7 & 69.6 & \textbf{70.4} & 8.8 & 13.9 & 38.6 & 75.1 & 75.3 & \textbf{79.8} \\
\multicolumn{1}{c|}{$y_t$=1} & 18.3 & 26.9 & 50.3 & 82.0 & 82.0 & \textbf{82.4} & 18.3 & 28.6 & 51.2 & 85.7 & 86.1 & \textbf{87.8} \\
\multicolumn{1}{c|}{$y_t$=2} & 27.8 & 37.4 & 45.8 & 87.6 & 87.6 & \textbf{87.9} & 29.2 & 38.7 & 46.4 & 89.1 & 89.1 & \textbf{92.3} \\
\multicolumn{1}{c|}{$y_t$=3} & 12.6 & 18.9 & 45.0 & 75.1 & 74.7 & \textbf{75.6} & 12.6 & 20.4 & 54.0 & 84.1 & 84.1 & \textbf{87.4} \\
\multicolumn{1}{c|}{$y_t$=4} & 8.6 & 15.3 & 36.7 & 74.3 & 74.2 & \textbf{75.3} & 9.0 & 16.3 & 38.0 & 81.0 & 81.5 & \textbf{85.8} \\
\multicolumn{1}{c|}{$y_t$=5} & 3.5 & 7.3 & 25.9 & 57.1 & 56.7 & \textbf{58.2} & 3.8 & 8.4 & 27.1 & 71.8 & 71.8 & \textbf{75.9} \\
\multicolumn{1}{c|}{$y_t$=6} & 19.9 & 27.3 & 47.1 & 81.7 & 81.6 & \textbf{82.1} & 18.3 & 26.4 & 50.8 & 84.1 & 84.2 & \textbf{87.5} \\ \midrule
\multirow{2}{*}{Cora} & \multicolumn{6}{c}{TAGCN} & \multicolumn{6}{c}{GCNII} \\ \cmidrule(l){2-13} 
 & Clean & Random & MostAttr & AFGSM & GB-FGSM & G$^2$-SNIA & Clean & Random & MostAttr & AFGSM & GB-FGSM & G$^2$-SNIA \\ \midrule
\multicolumn{1}{c|}{$y_t$=0} & 9.1 & 14.9 & 28.2 & 57.5 & 57.0 & \textbf{62.6} & 9.7 & 12.7 & 25.5 & 33.7 & 33.2 & \textbf{41.7} \\
\multicolumn{1}{c|}{$y_t$=1} & 17.5 & 20.9 & 33.3 & 70.9 & 71.0 & \textbf{75.9} & 18.1 & 23.0 & 48.8 & 66.8 & 67.8 & \textbf{72.4} \\
\multicolumn{1}{c|}{$y_t$=2} & 30.0 & 39.8 & 44.1 & 87.1 & 87.2 & \textbf{91.7} & 29.5 & 36.0 & 45.1 & 78.2 & 78.2 & \textbf{83.5} \\
\multicolumn{1}{c|}{$y_t$=3} & 12.7 & 18.4 & 39.6 & 67.9 & 68.3 & \textbf{73.6} & 12.7 & 18.0 & 44.6 & 58.6 & 58.5 & \textbf{67.7} \\
\multicolumn{1}{c|}{$y_t$=4} & 8.8 & 14.7 & 30.8 & 64.4 & 64.3 & \textbf{70.4} & 8.4 & 11.6 & 26.7 & 31.7 & 31.5 & \textbf{41.8} \\
\multicolumn{1}{c|}{$y_t$=5} & 5.2 & 10.0 & 22.1 & 53.2 & 52.9 & \textbf{58.8} & 3.7 & 5.6 & 11.5 & 14.0 & 13.8 & \textbf{16.8} \\
\multicolumn{1}{c|}{$y_t$=6} & 16.7 & 23.3 & 39.4 & 74.6 & 74.6 & \textbf{80.2} & 17.9 & 22.6 & 31.1 & 48.2 & 48.2 & \textbf{59.5} \\ \midrule
\midrule
\multirow{2}{*}{Citeseer} & \multicolumn{6}{c}{GCN} & \multicolumn{6}{c}{SGCN} \\ \cmidrule(l){2-13} 
 & Clean & Random & MostAttr & AFGSM & GB-FGSM & G$^2$-SNIA & Clean & Random & MostAttr & AFGSM & GB-FGSM & G$^2$-SNIA \\ \midrule
\multicolumn{1}{c|}{$y_t$=0} & 5.8 & 19.2 & 18.3 & 78.8 & 79.1 & \textbf{82.0} & 5.4 & 16.0 & 16.7 & 82.8 & 82.9 & \textbf{87.5} \\
\multicolumn{1}{c|}{$y_t$=1} & 24.3 & 33.0 & 46.4 & 78.6 & 78.6 & \textbf{82.0} & 24.6 & 31.6 & 46.6 & 81.9 & 82 & \textbf{84.9} \\
\multicolumn{1}{c|}{$y_t$=2} & 15.9 & 25.1 & 35.6 & 78.0 & 78.2 & \textbf{81.2} & 18.7 & 32.5 & 49.0 & 86.4 & 86.5 & \textbf{90.5} \\
\multicolumn{1}{c|}{$y_t$=3} & 13.1 & 18.6 & 40.3 & 72.5 & 72.7 & \textbf{77.0} & 12.9 & 22.0 & 48.7 & 79.9 & 79.8 & \textbf{85.5} \\
\multicolumn{1}{c|}{$y_t$=4} & 30.6 & 43.0 & 56.9 & 87.7 & 87.7 & \textbf{90.2} & 28.2 & 37.6 & 50.8 & 86.1 & 86.1 & \textbf{90.3} \\
\multicolumn{1}{c|}{$y_t$=5} & 10.3 & 16.7 & 25.2 & 73.3 & 73.5 & \textbf{76.3} & 10.2 & 17.0 & 26.9 & 81.4 & 81.5 & \textbf{87.2} \\ \midrule
\multirow{2}{*}{Citeseer} & \multicolumn{6}{c}{TAGCN} & \multicolumn{6}{c}{GCNII} \\ \cmidrule(l){2-13} 
 & Clean & Random & MostAttr & AFGSM & GB-FGSM & G$^2$-SNIA & Clean & Random & MostAttr & AFGSM & GB-FGSM & G$^2$-SNIA \\ \midrule
\multicolumn{1}{c|}{$y_t$=0} & 6.8 & 16.0 & 27.3 & 70.6 & 70.5 & \textbf{81.1} & 2.9 & 4.0 & 7.9 & 12.2 & 12.4 & \textbf{33.6} \\
\multicolumn{1}{c|}{$y_t$=1} & 26.9 & 33.8 & 42.7 & 76.5 & 76.8 & \textbf{84.3} & 28.0 & 35.9 & 48.5 & 71.8 & 72.1 & \textbf{80.5} \\
\multicolumn{1}{c|}{$y_t$=2} & 15.2 & 21.6 & 27.3 & 70.4 & 71.2 & \textbf{78.9} & 17.8 & 24.3 & 36.5 & 48.5 & 48.6 & \textbf{61.1} \\
\multicolumn{1}{c|}{$y_t$=3} & 11.6 & 21.6 & 47.1 & 81.9 & 82.0 & \textbf{89.4} & 12.2 & 19.7 & 44.9 & 51.6 & 50.4 & \textbf{67.7} \\
\multicolumn{1}{c|}{$y_t$=4} & 24.8 & 28.8 & 30.7 & 73.2 & 73.5 & \textbf{82.3} & 31.5 & 40.5 & 47.9 & 81.2 & 81.5 & \textbf{92.1} \\
\multicolumn{1}{c|}{$y_t$=5} & 14.7 & 23.1 & 28.2 & 76.0 & 76.4 & \textbf{84.3} & 7.6 & 9.8 & 16.1 & 17.5 & 17.0 & \textbf{37.8} \\ \midrule
\midrule
\multirow{2}{*}{DBLP} & \multicolumn{6}{c}{GCN} & \multicolumn{6}{c}{SGCN} \\ \cmidrule(l){2-13} 
 & Clean & Random & MostAttr & AFGSM & GB-FGSM & G$^2$-SNIA & Clean & Random & MostAttr & AFGSM & GB-FGSM & G$^2$-SNIA \\ \midrule
\multicolumn{1}{c|}{$y_t$=0} & 49.0 & 61.6 & 92.6 & 95.1 & 95.1 & \textbf{95.2} & 48.3 & 57.5 & 91.7 & 96.2 & 96.2 & \textbf{96.3} \\
\multicolumn{1}{c|}{$y_t$=1} & 29.5 & 38.8 & 51.3 & \textbf{78.3} & \textbf{78.3} & \textbf{78.3} & 30.6 & 38.6 & 48.1 & 85.5 & 85.5 & \textbf{86.1} \\
\multicolumn{1}{c|}{$y_t$=2} & 11.0 & 15.8 & 54.4 & \textbf{64.3} & 64.2 & 64.1 & 10.6 & 16.6 & 63.6 & 80.0 & 80.0 & \textbf{80.8} \\
\multicolumn{1}{c|}{$y_t$=3} & 10.5 & 18.0 & 53.9 & 65.6 & 65.3 & \textbf{66.9} & 10.5 & 19.6 & 60.6 & 75.8 & 75.5 & \textbf{80.5} \\ \midrule
\multirow{2}{*}{DBLP} & \multicolumn{6}{c}{TAGCN} & \multicolumn{6}{c}{GCNII} \\ \cmidrule(l){2-13} 
 & Clean & Random & MostAttr & AFGSM & GB-FGSM & G$^2$-SNIA & Clean & Random & MostAttr & AFGSM & GB-FGSM & G$^2$-SNIA \\ \midrule
\multicolumn{1}{c|}{$y_t$=0} & 47.5 & 54.9 & 89.0 & 93.3 & 93.3 & \textbf{94.1} & 46.1 & 55.4 & 90.3 & 94.8 & 94.8 & \textbf{96.4} \\
\multicolumn{1}{c|}{$y_t$=1} & 27.7 & 34.8 & 40.2 & 73.6 & 73.7 & \textbf{76.5} & 28.7 & 33.9 & 43.6 & 69.5 & 69.6 & \textbf{74.6} \\
\multicolumn{1}{c|}{$y_t$=2} & 10.8 & 13.7 & 33.0 & 40.2 & 40.6 & \textbf{49.1} & 11.6 & 18.5 & 49.9 & 63.6 & 63.3 & \textbf{65.1} \\
\multicolumn{1}{c|}{$y_t$=3} & 14.0 & 18.3 & 32.5 & 37.6 & 37.3 & \textbf{60.1}& 13.6 & 17.2 & 36.7 & 47.8 & 47.6 & \textbf{55.8} \\ \bottomrule
\end{tabular*}
\end{table*}
\subsection{Effectiveness Comparison}
%\subsubsection{Evaluation of effevtiveness.}
To address \textbf{RQ1}, we first evaluate the performance of our proposed method, G$^2$-SNIA, in comparison with four baseline methods in the black-box evasion setting. The classification results of the original graph, denoted as Clean, are also included as a lower bound. Table \ref{black} presents the attack success rate for different targeted labels. As the results in the table show, all methods, including Random, are capable of performing label specificity attack against the four victim GNNs. Random has a modest attack effect, indicating that nodes belonging to the targeted label can spread perturbations to the target node via the aggregation process of GNNs. MostAtrr performs better than Random, which indicates the injected node has better attack performance than the node in the original graph even when generated based solely on statistics. Regarding the state-of-the-art attack baselines, both AFGSM and GB-FGSM demonstrate the best performance among all baselines, demonstrating that gradient-based intentional perturbations can effectively mislead classification results made by victim models for target nodes. Our proposed method either outperforms or performs on par with all baselines across attacks against all victim models. 
\begin{table*}[!ht]
  \centering
  \caption{The Average attack Success Rate Comparison in different attack scenario}
  \label{white}
  \centering
  \begin{tabular}{@{}l|cccc|cccc|cccc@{}}
  \toprule
   & \multicolumn{4}{c|}{Cora} & \multicolumn{4}{c|}{Citeseer} & \multicolumn{4}{c}{DBLP} \\ \midrule
   & GCN & SGCN & TAGCN & GCNII & GCN & SGCN & TAGCN & GCNII & GCN & SGCN & TAGCN & GCNII \\ \midrule
  AFGSM (White-box) & 75.07 & 85.26 & 71.56 & 47.39 & 80.23 & 87.63 & 78.73 & 51.28 & 75.45 & 85.55 & 53.45 & 60.25 \\
  GB-FGSM (White-box) & \textbf{76.41} & \textbf{85.36} & \textbf{73.69} & 54.76 & \textbf{81.67} & \textbf{87.67} & \textbf{83.63} & \textbf{62.43} & \textbf{76.38} & \textbf{85.98} & 68.10 & 72.18 \\
  GB-FGSM (Black-box) & 75.20 & 81.73 & 67.90 & 47.31 & 78.30 & 83.13 & 75.07 & 47.00 & 75.72 & 84.30 & 61.23 & 68.82 \\ 
  AFGSM (Black-box) & 75.36 & 81.56 & 67.94 & 47.31 & 78.15 & 83.08 & 74.77 & 47.13 & 75.82 & 84.38 & 61.17 & 68.92 \\ 
  G$^2$-SNIA  & 75.99 & 85.21 & 73.31 & \textbf{54.77} & 81.45 & 87.65 & 83.38 & 62.13 & 76.12 & 85.93 & \textbf{69.95} & \textbf{72.97} \\ 
  \bottomrule
  \end{tabular}
  \end{table*}
\begin{figure*}[!ht]
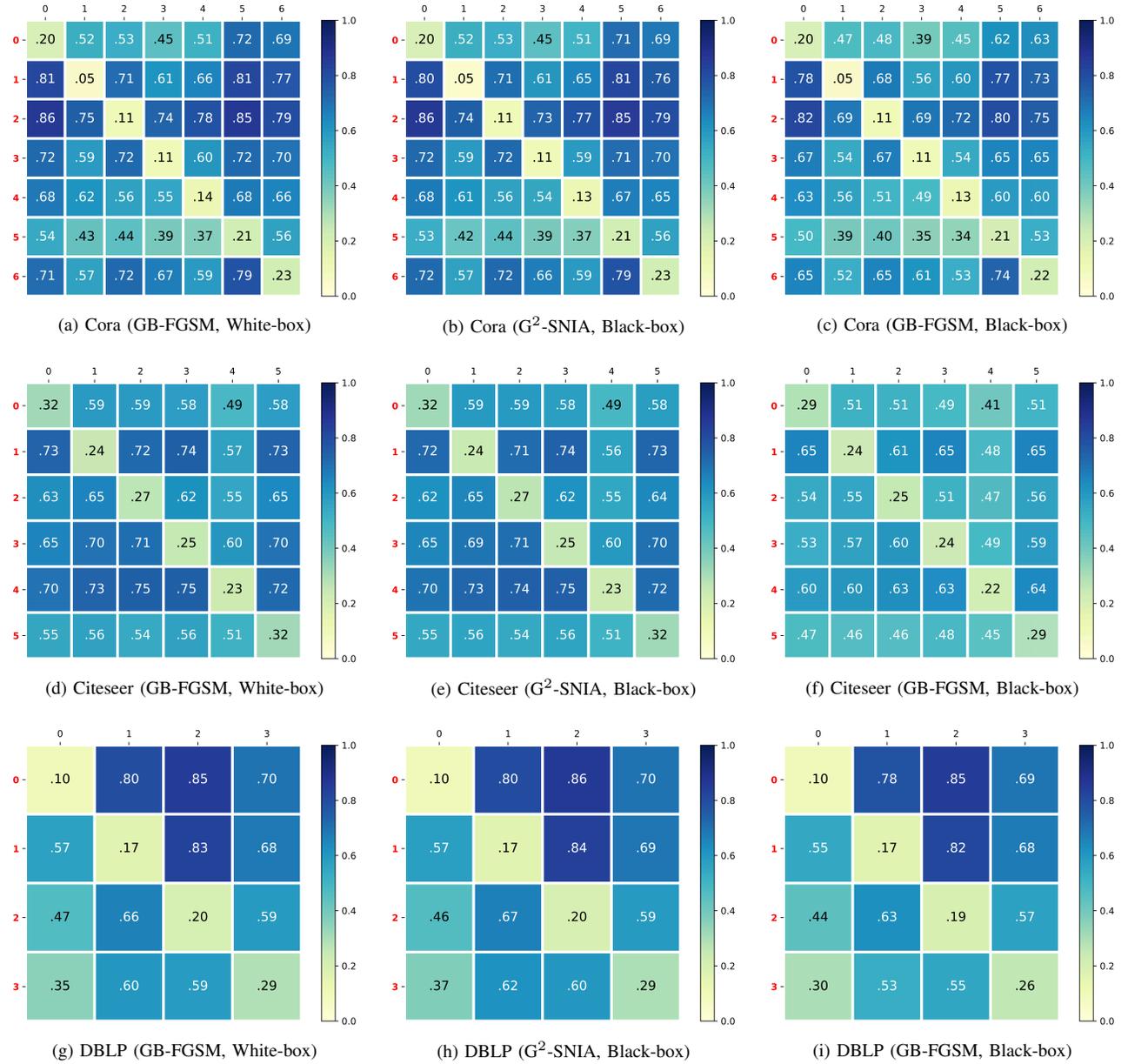

    \centering
    \subfloat[Cora (GB-FGSM, White-box)]{\includegraphics[width=0.3\textwidth]{heatmap/cora_gb_fgsm.pdf}%
    \label{cora_gb_fgsm}}
    \quad
    \subfloat[Cora (G$^2$-SNIA, Black-box)]{\includegraphics[width=0.3\textwidth]{heatmap/cora_gf_snia.pdf}%
    \label{cora_gf_snia}}
    \quad
    \subfloat[Cora (GB-FGSM, Black-box)]{\includegraphics[width=0.3\textwidth]{heatmap/black_cora_gb_fgsm.pdf}%
    \label{black_cora_gb_fgsm}}
    \\
    \subfloat[Citeseer (GB-FGSM, White-box)]{\includegraphics[width=0.3\textwidth]{heatmap/citeseer_gb_fgsm.pdf}%
    \label{citeseer_gb_fgsm}}
    \quad
    \subfloat[Citeseer (G$^2$-SNIA, Black-box)]{\includegraphics[width=0.3\textwidth]{heatmap/citeseer_gf_snia.pdf}%
    \label{citeseer_gf_snia}}
    \quad
    \subfloat[Citeseer (GB-FGSM, Black-box)]{\includegraphics[width=0.3\textwidth]{heatmap/black_citeseer_gb_fgsm.pdf}%
    \label{black_citeseer_gb_fgsm}}
    \\
    \subfloat[DBLP (GB-FGSM, White-box)]{\includegraphics[width=0.3\textwidth]{heatmap/dblp_gb_fgsm.pdf}%
    \label{dblp_gb_fgsm}}
    \quad
    \subfloat[DBLP (G$^2$-SNIA, Black-box)]{\includegraphics[width=0.3\textwidth]{heatmap/dblp_gf_snia.pdf}%
    \label{dblp_gf_snia}}
    \quad
    \subfloat[DBLP (GB-FGSM, Black-box)]{\includegraphics[width=0.3\textwidth]{heatmap/black_dblp_gb_fgsm.pdf}%
    \label{black_dblp_gb_fgsm}}
    \caption{Changes in average targeted label confidence on all datasets and victim models. Darker color suggests larger performance promotion.}
    \label{heatmap}
    \end{figure*}
In all datasets and models, G$^2$-SNIA approximately surpasses the best baseline by 5\% averagely. In some cases, it even reaches about 20\% improvement. These indicate that although AFGSM and GB-FGSM rely on the maximum gradient strategy to generate perturbations, their solutions might represent a good local optimum for surrogate models. However, when transferred to victim models, these strategies are insufficient. Without requiring gradient information of victim models, G$^2$-SNIA accurately identifies the perturbations with the most significant impact on the results, thereby thoroughly surpassing the attack performance generated by transfer attack leveraging surrogate gradients.
We also observe that different victim models exhibit varying sensitivities to perturbations. Generally, GCN and SGCN are more susceptible to perturbations, while GCNII exhibits greater robustness. For example, in the Cora dataset, G$^2$-SNIA achieves a 75.99\% average success rate on the GCN model but only 54.77\% on GCNII. Meanwhile, We note that some labels are more challenging to execute label specificity attack under the same conditions. For instance, in the Cora dataset and GCNII model, G$^2$-SNIA achieves an 83.5\% success rate for $y_t=2$ but only 16.8\% for $y_t=5$. This discrepancy may be attributed to the distribution of node labels in the training set and the architectures of classification GNN models.
\begin{figure*}[!ht]
\centering
\subfloat[Cora]{\includegraphics[width=0.3\textwidth]{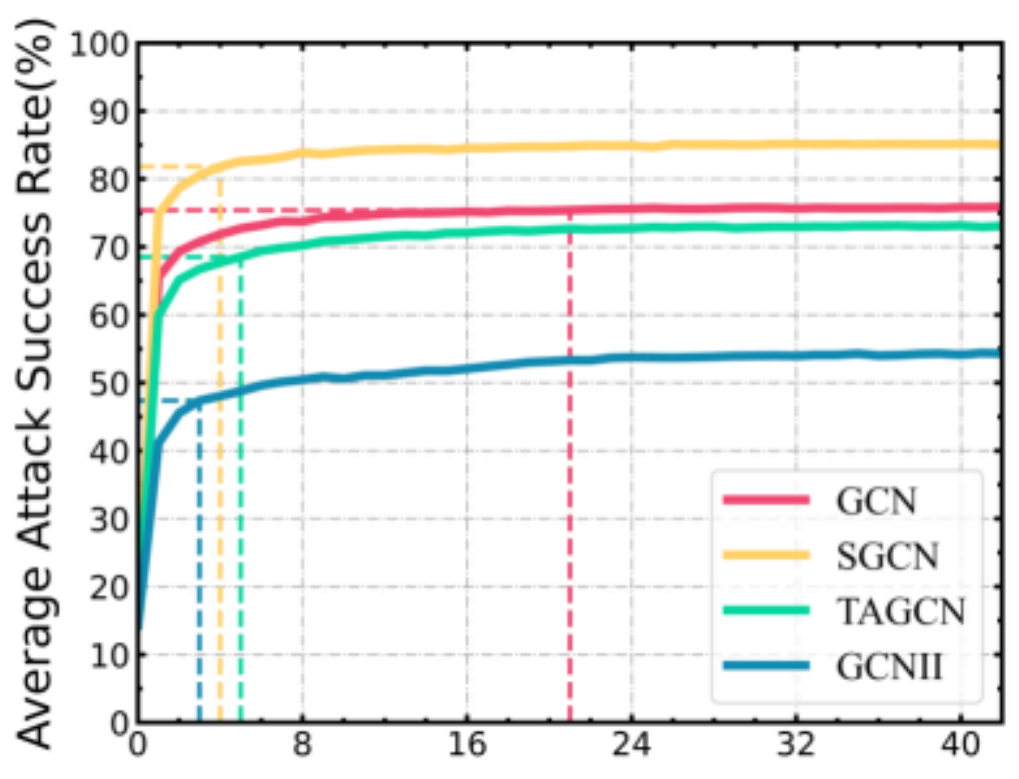}%
\label{cora}}
\quad
\subfloat[Citeseer]{\includegraphics[width=0.3\textwidth]{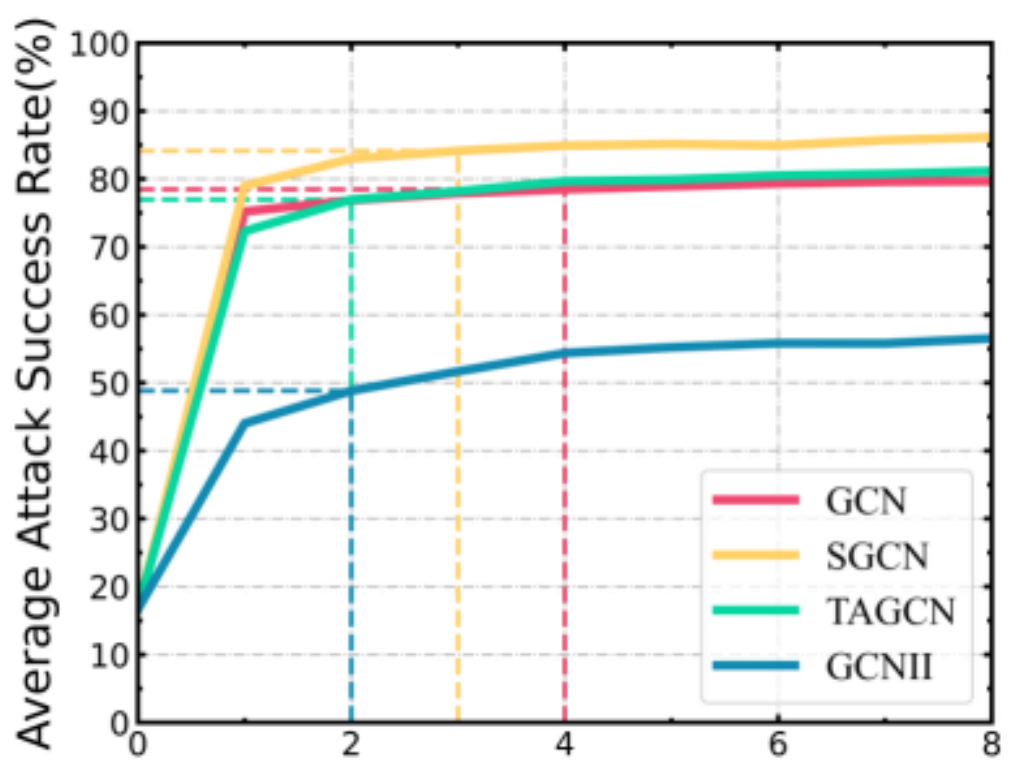}%
\label{citeseer}}
\quad
\subfloat[DBLP]{\includegraphics[width=0.3\textwidth]{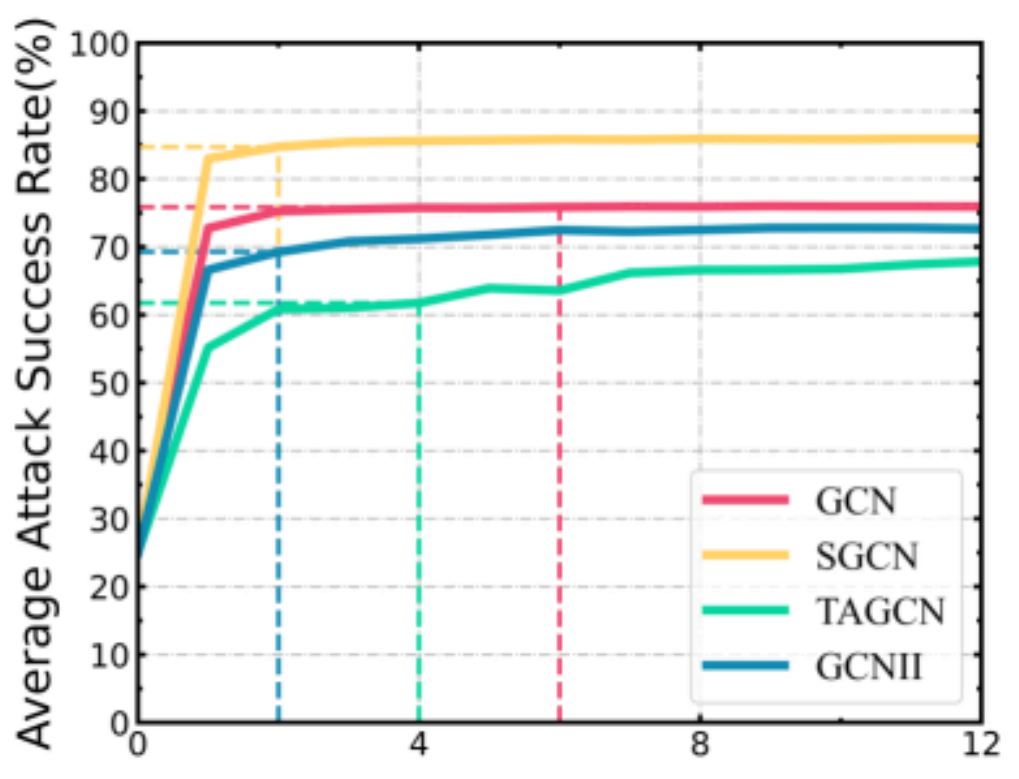}%
\label{dblp}}
\caption{Change of average attack success rate with the amount of exploration.}
\label{curve}
\end{figure*}
\begin{figure*}[!ht]
\centering
\subfloat[Cora(GCN)]{\includegraphics[width=0.22\textwidth]{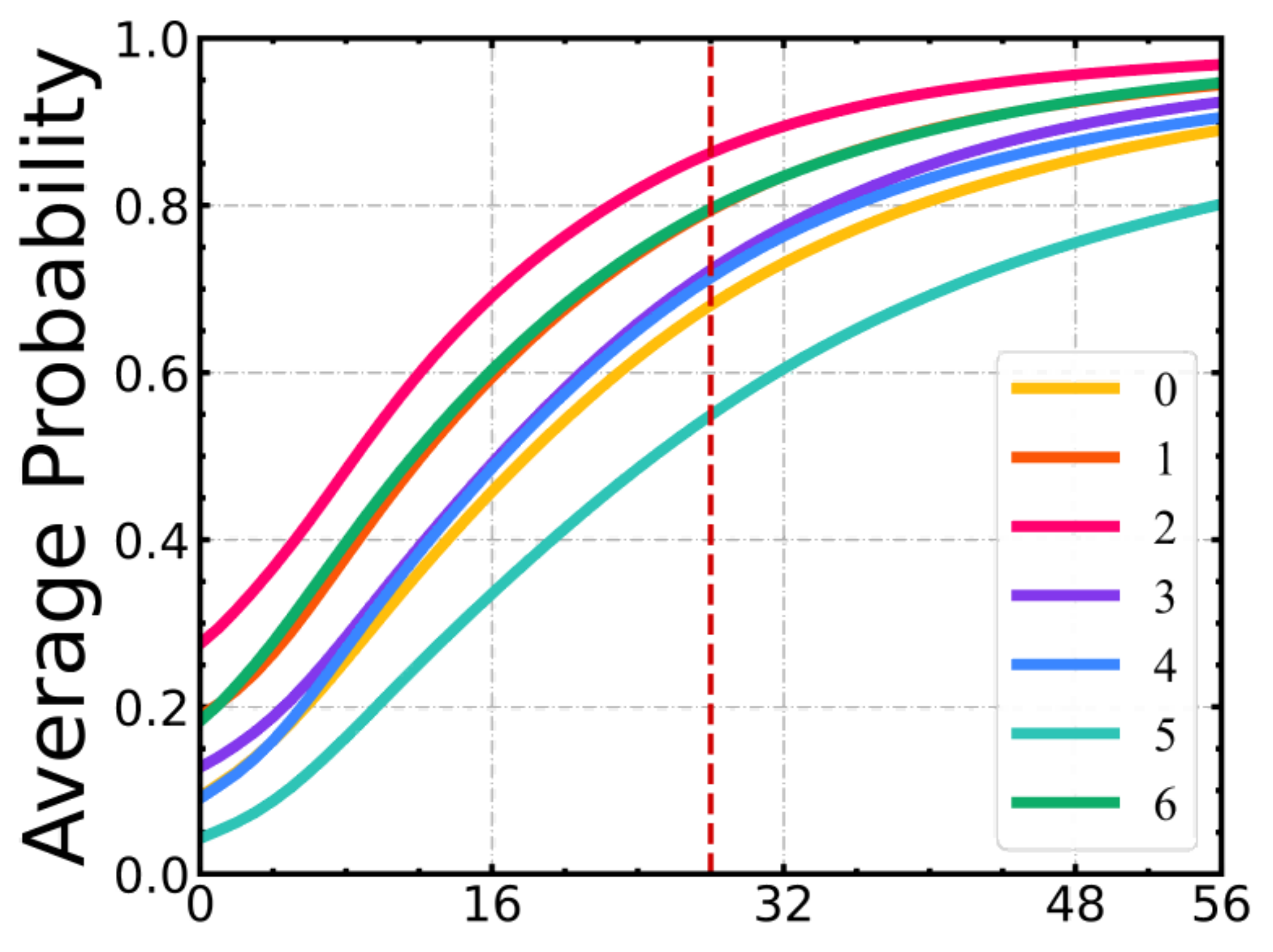}%
\label{cora_gcn}}
\quad
\subfloat[Cora(SGCN)]{\includegraphics[width=0.22\textwidth]{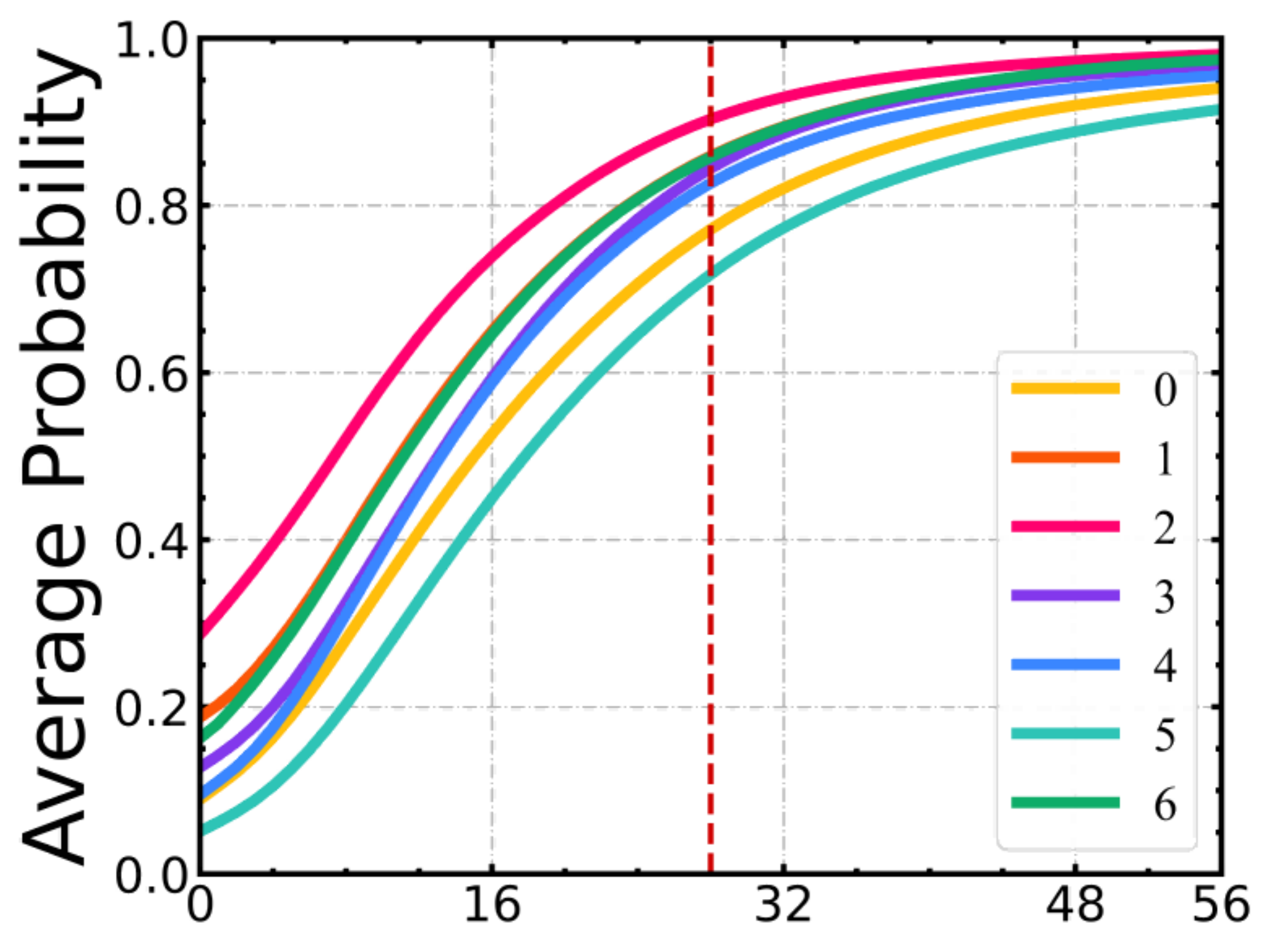}%
\label{cora_sgcn}}
\quad
\subfloat[Cora(TAGCN)]{\includegraphics[width=0.22\textwidth]{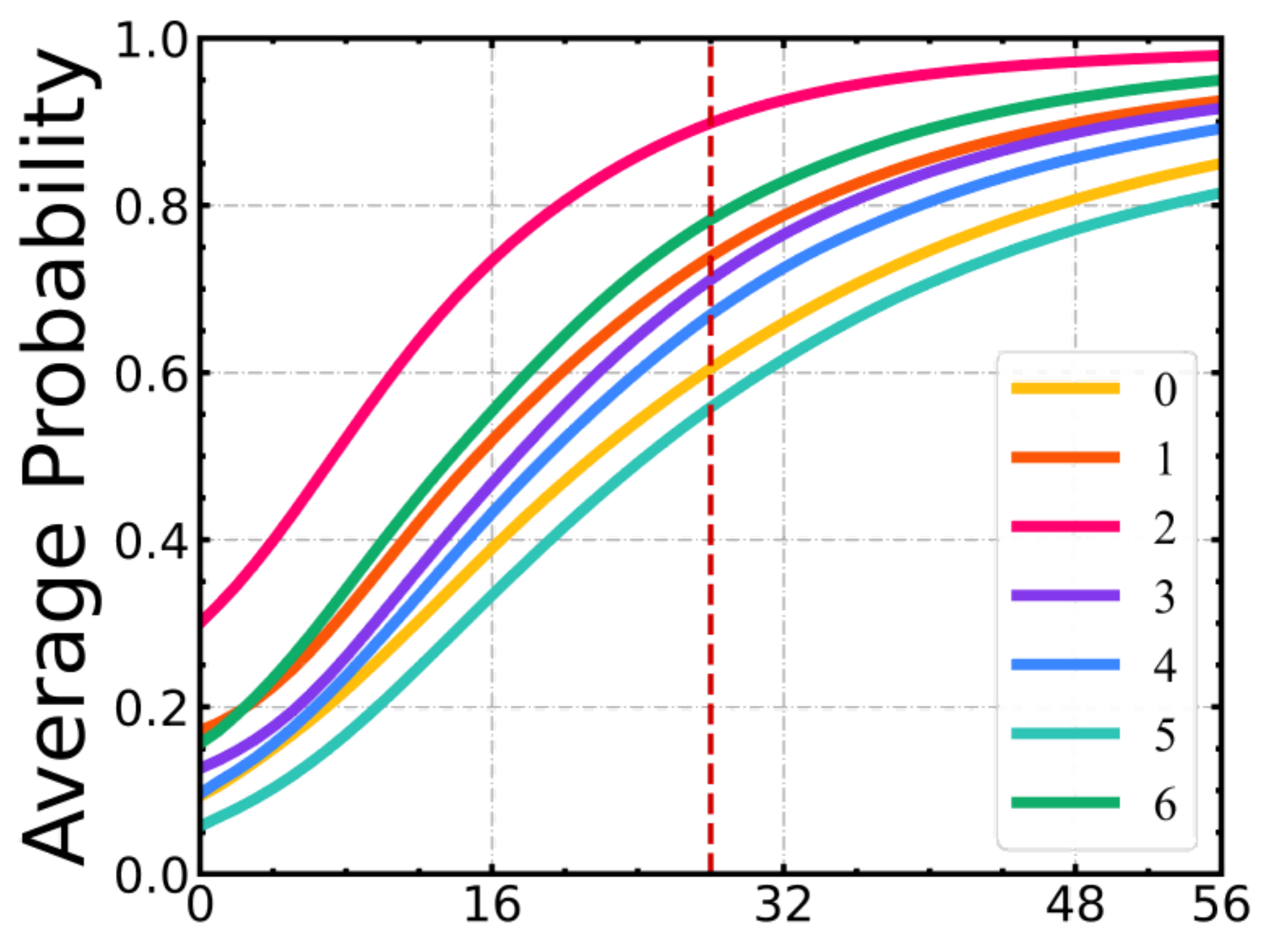}%
\label{cora_tagcn}}
\quad
\subfloat[Cora(GCNII)]{\includegraphics[width=0.22\textwidth]{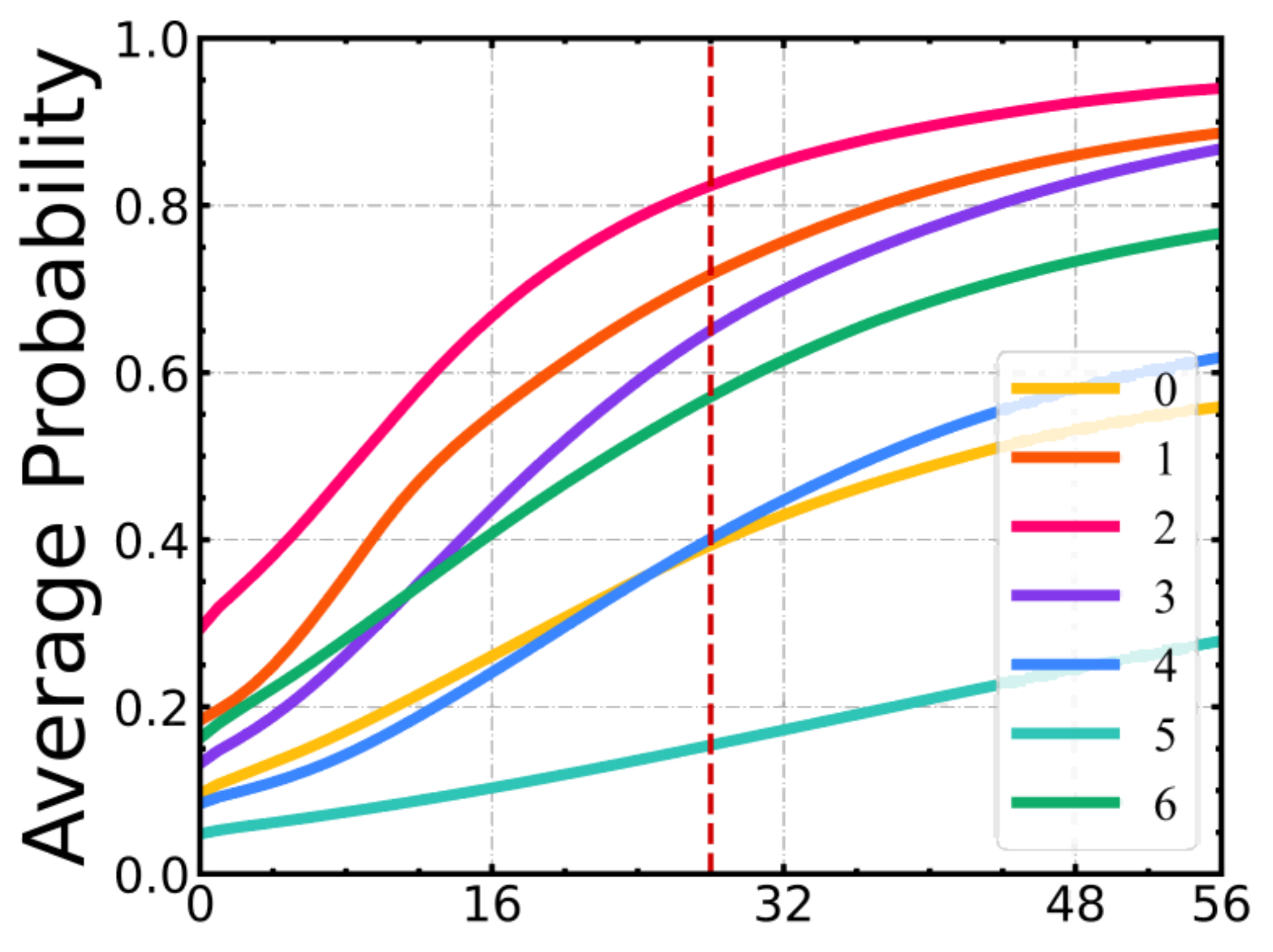}%
\label{cora_gcnii}}
\\
\subfloat[Citeseer(GCN)]{\includegraphics[width=0.22\textwidth]{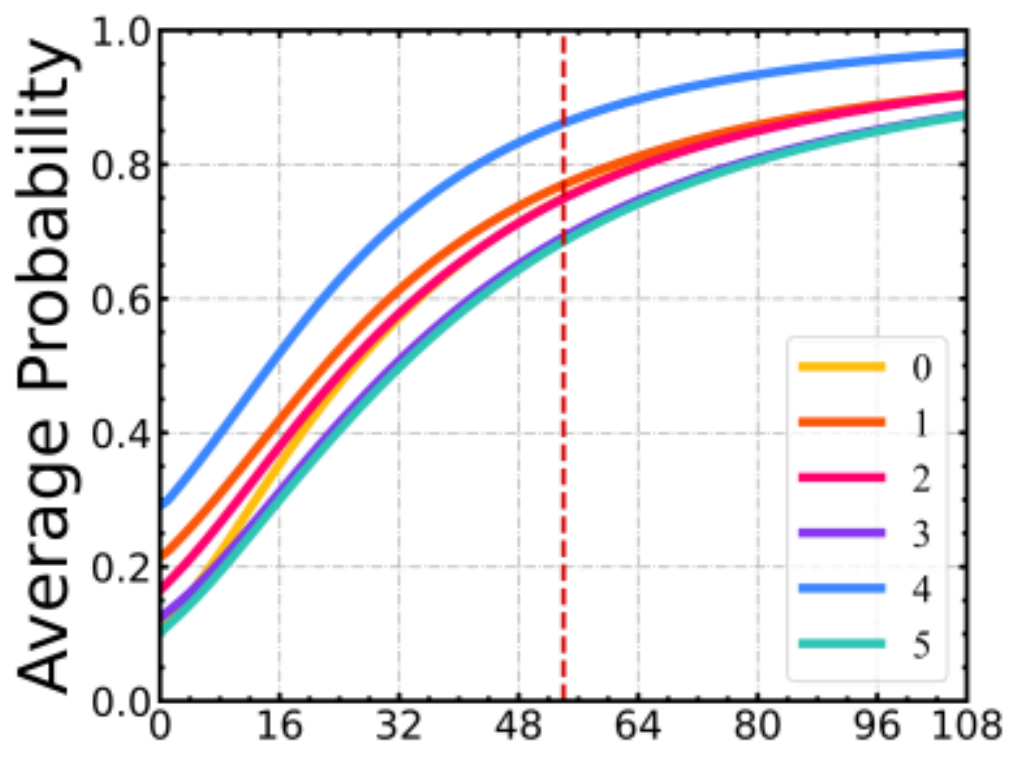}%
\label{citeseer_gcn}}
\quad
\subfloat[Citeseer(SGCN)]{\includegraphics[width=0.22\textwidth]{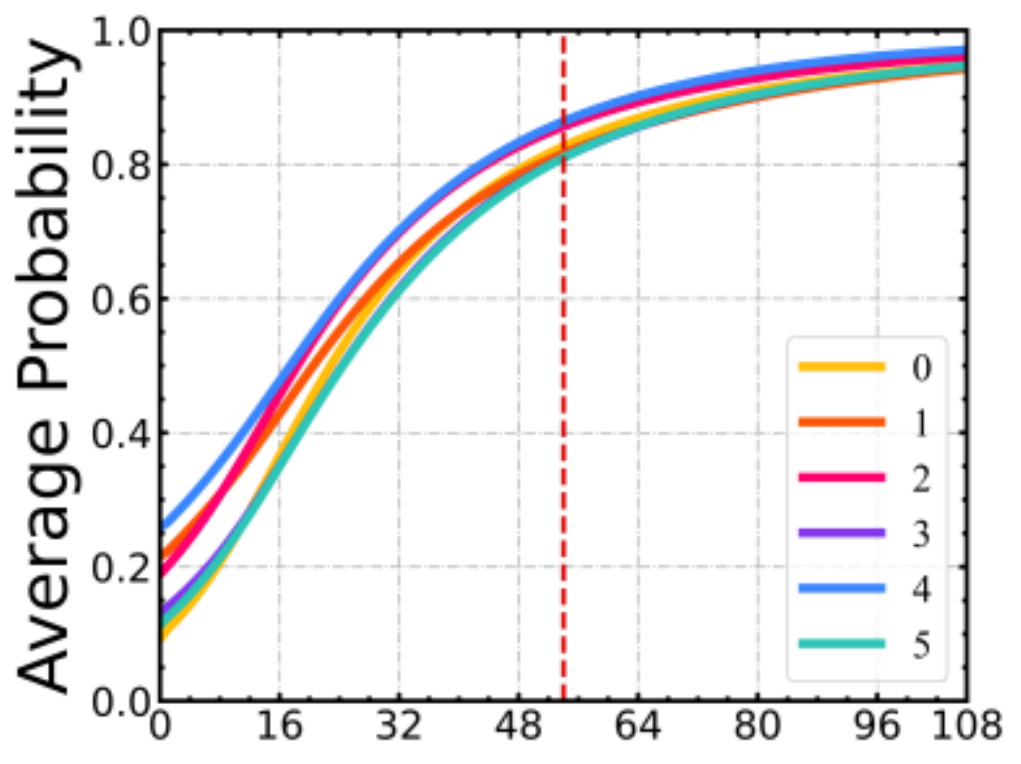}%
\label{citeseer_sgcn}}
\quad
\subfloat[Citeseer(TAGCN)]{\includegraphics[width=0.22\textwidth]{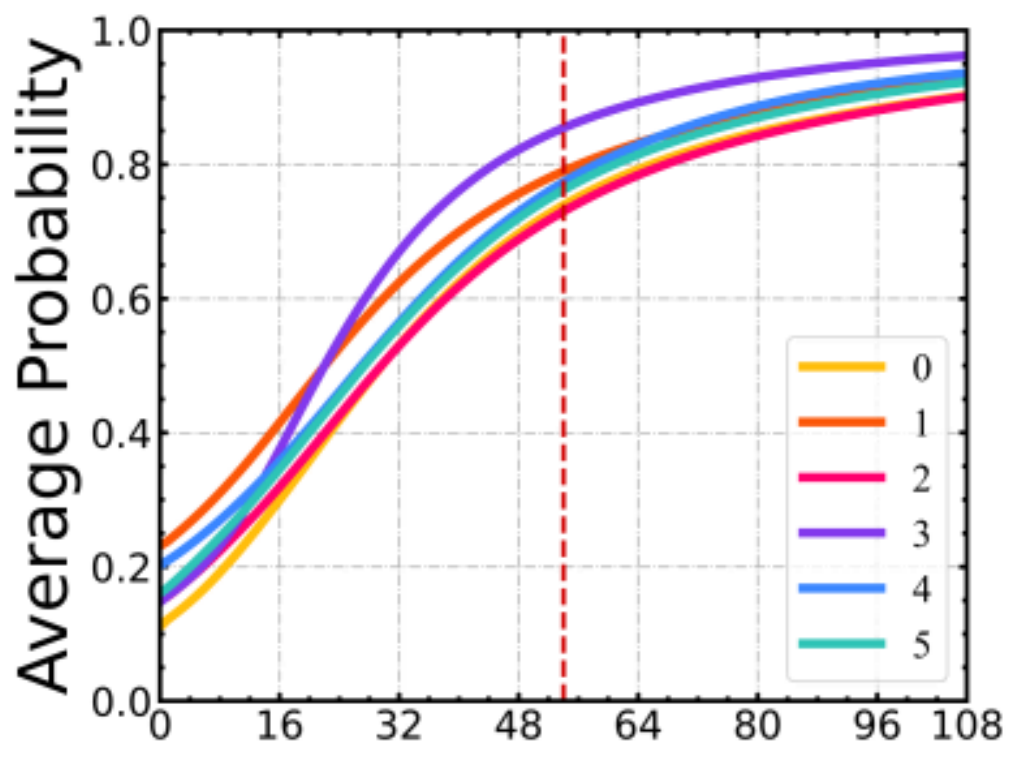}%
\label{citeseer_tagcn}}
\quad
\subfloat[Citeseer(GCNII)]{\includegraphics[width=0.22\textwidth]{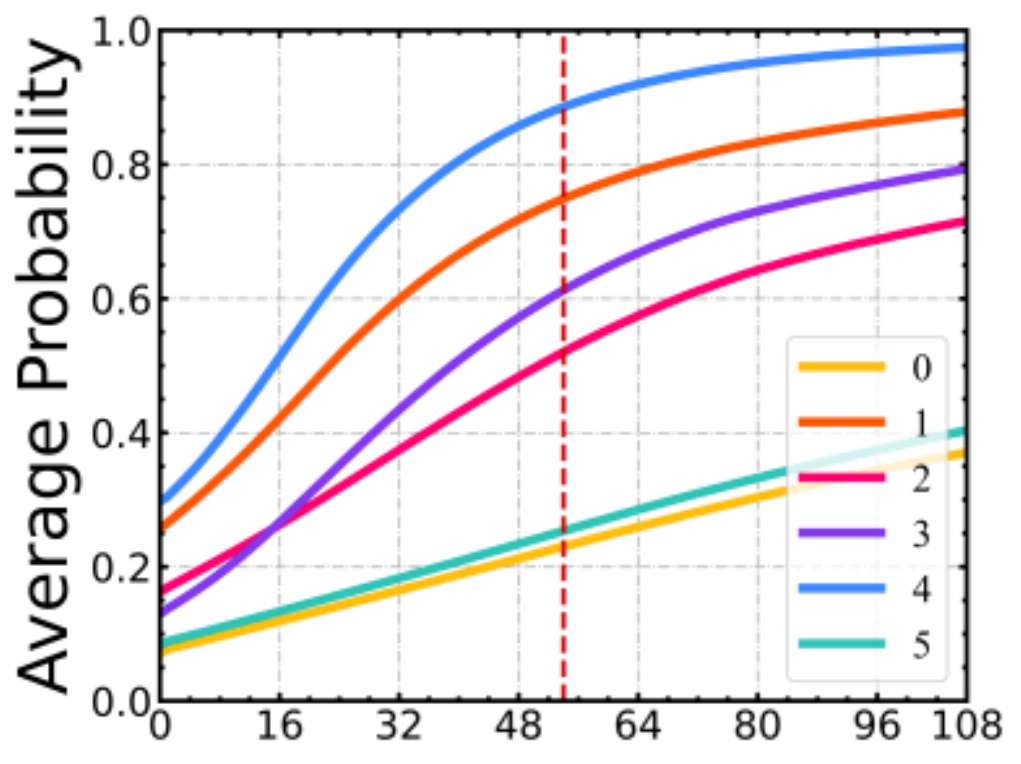}%
\label{citeseer_gcnii}}
\\
\subfloat[DBLP(GCN)]{\includegraphics[width=0.22\textwidth]{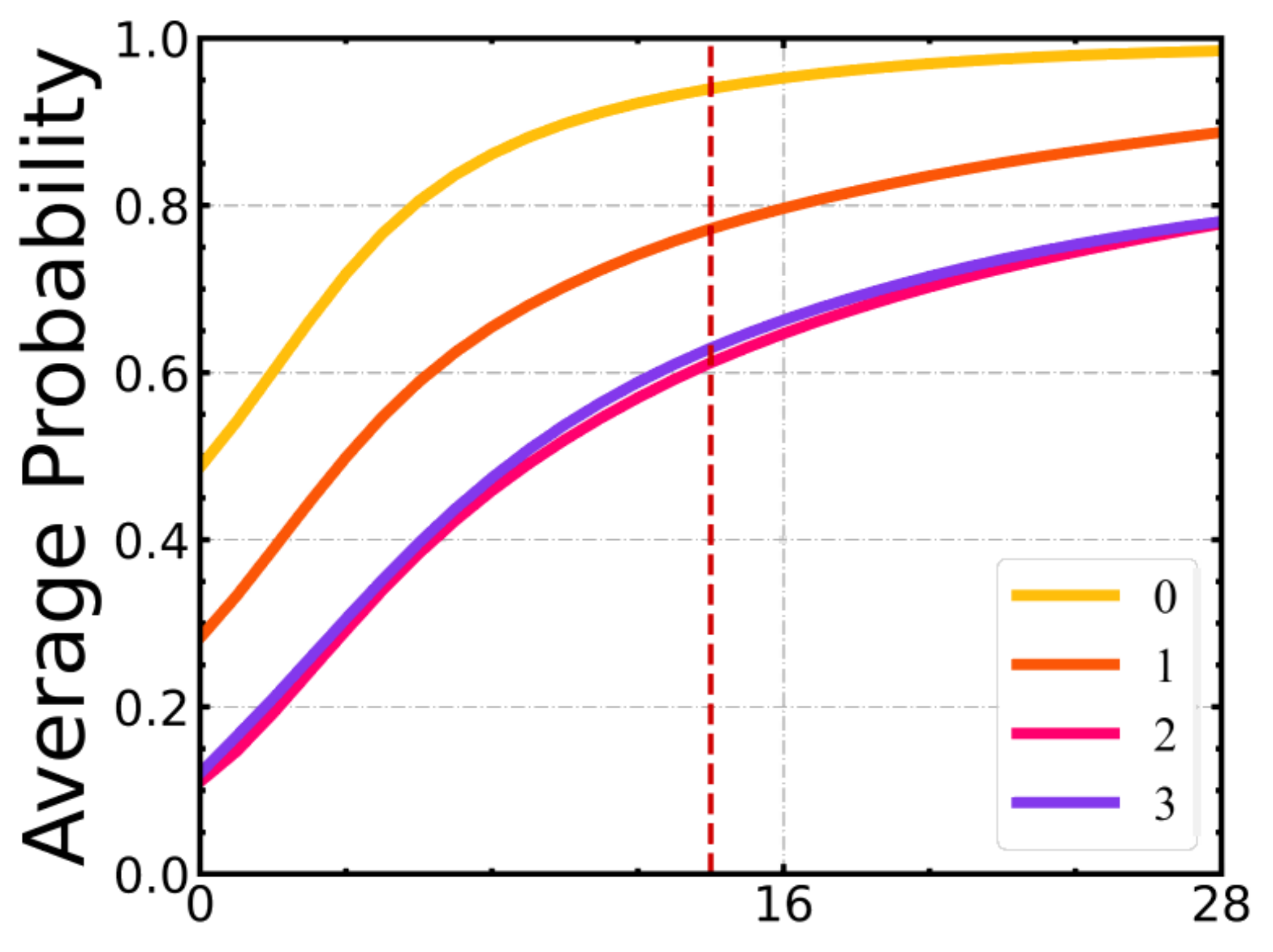}%
\label{dblp_gcn}}
\quad
\subfloat[DBLP(SGCN)]{\includegraphics[width=0.22\textwidth]{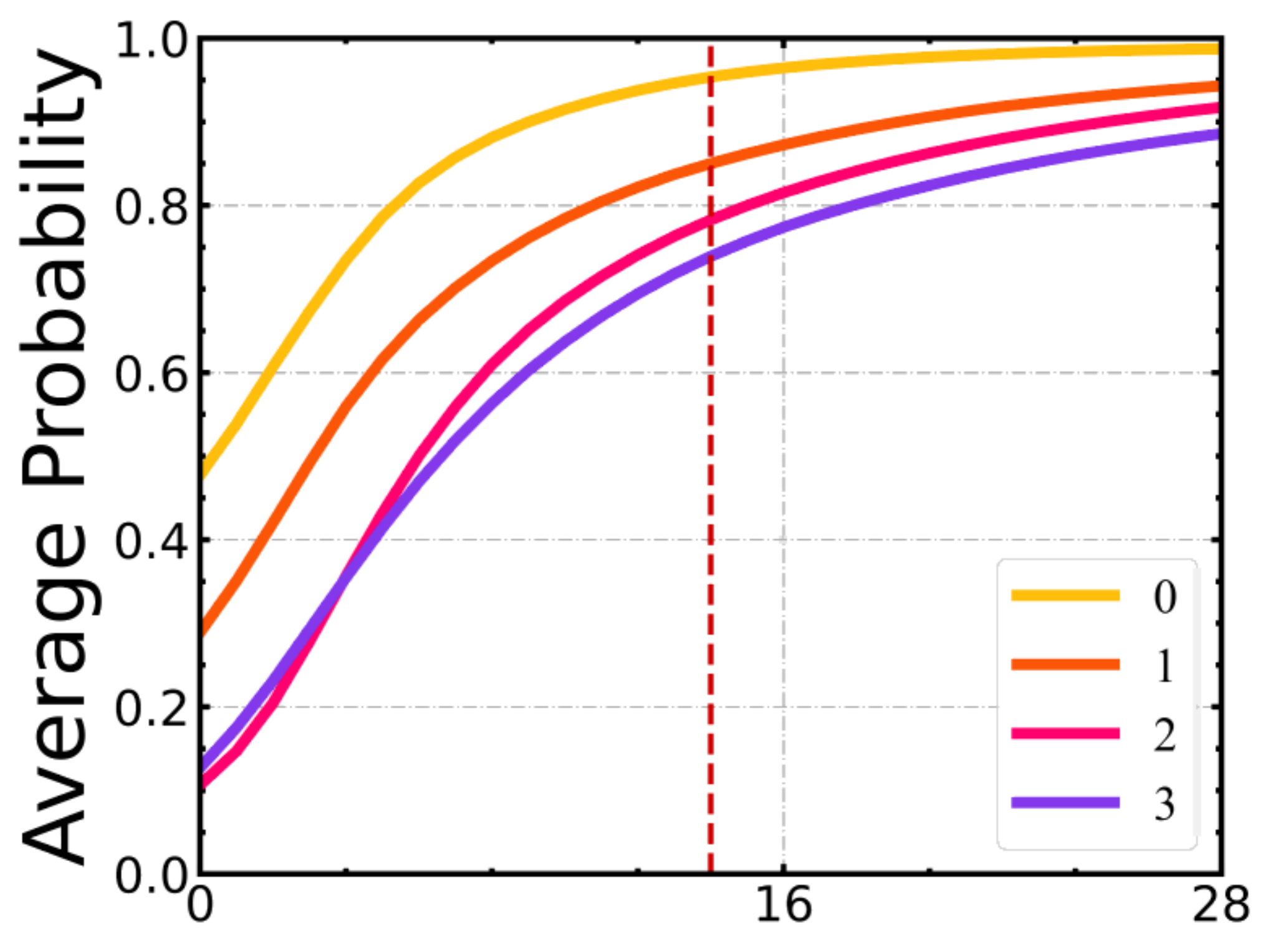}%
\label{dblp_sgcn}}
\quad
\subfloat[DBLP(TAGCN)]{\includegraphics[width=0.22\textwidth]{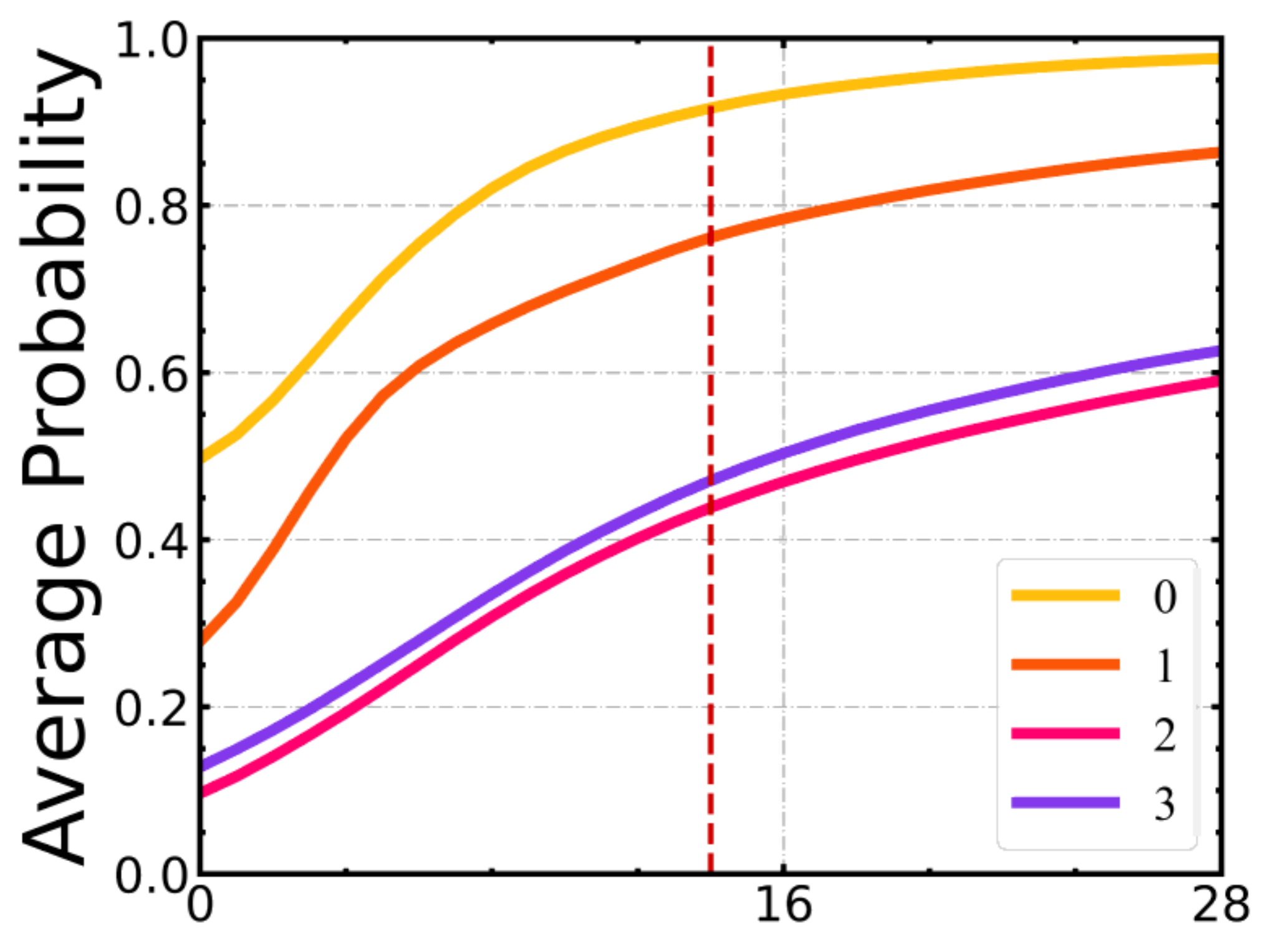}%
\label{dblp_tagcn}}
\quad
\subfloat[DBLP(GCNII)]{\includegraphics[width=0.22\textwidth]{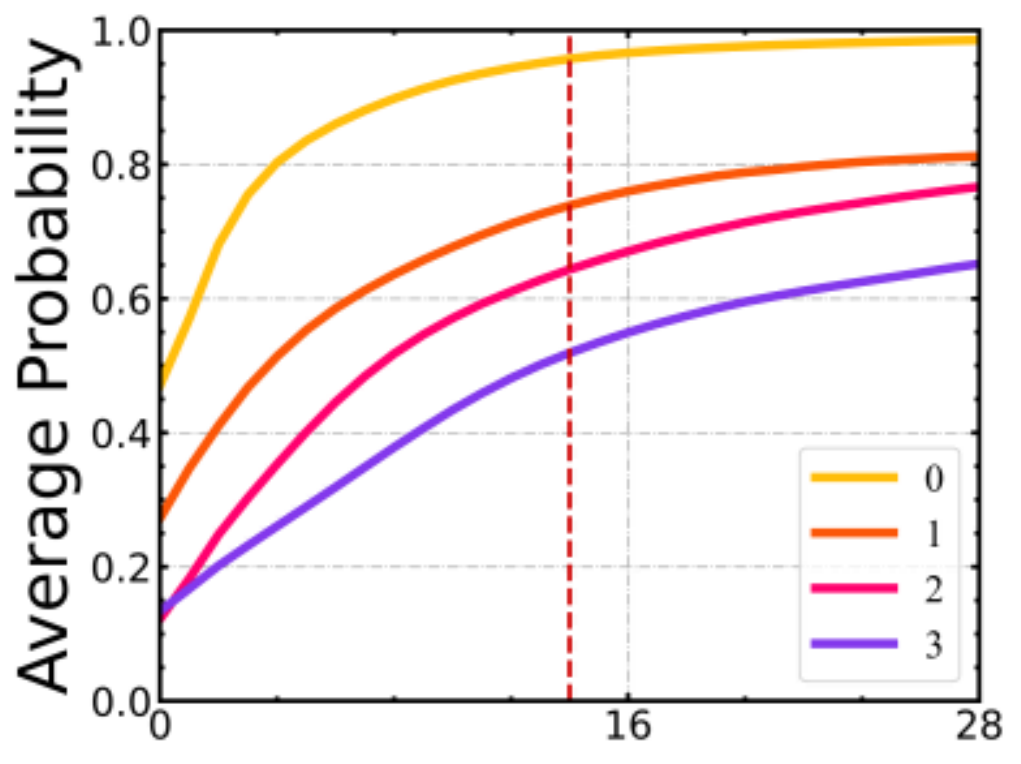}%
\label{dblp_gcnii}}
\caption{The trend of the average probability of attacked nodes belonging to the targeted label under all datasets and victim models.}
\label{line_graph}
\end{figure*}
% 这个地方设计得很精妙

Besides effectiveness comparison in the black-box setting, we also compare the solutions obtained by our method with the local optimal solutions generated by gradient-based baseline methods in the white-box setting. The results are presented in Table~\ref{white}. We observe that the baselines in the white-box setting exhibit similar performance and significantly outperform baselines in the black-box setting in most cases, which proves that the divergence between the surrogate architecture and the actual victim model results in attack performance decreasing.
G$^2$-SNIA outperforms AFGSM (white-box) in most cases and is slightly inferior to GB-FGSM (white-box), demonstrating that despite having limited information about the victim model, the solutions generated by our proposed method are comparable to the local optimal solutions obtained based on gradient-based greedy methods in the white-box setting, i.e., we are still able to obtain a good approximate solution for this NP-Hard problem. To further verify the performance of G$^2$-SNIA, we compare the attack performance of GB-FGSM (white-box), GB-FGSM (black-box) and  G$^2$-SNIA by measuring the change in the probability of target nodes being classified as the targeted label by victim models before and after the attack, defined as $\Delta \hat{\mathbf{Z}}_{v_t, y_t}= \hat{\mathbf{Z}}_{v_t, y_t} - \mathbf{Z}_{v_t, y_t}$. As shown in Fig.~\ref{heatmap}, the numbers on the x-axis represent the original classification labels of target nodes, i.e., the classification results of target nodes by victim models on the original graph and the red numbers on the y-axis represent the targeted labels. The value in each cell is denoted as $\frac{1}{|M||L_i|} \sum_{M} \sum_{v \in L_{i}} \Delta \hat{\mathbf{Z}}_{v,j}$, where $i$ is the original label, $j$ is the targeted label, $M$ is the set of victim models. The results in the figure show that G$^2$-SNIA is almost as effective as the state-of-the-art white-box attack method and outperforms the baseline in the black-box setting. Interestingly, the data on the diagonal in the figure suggests that single node injection label specificity attack can also be utilized as a method to enhance node classification. 
\subsection{Efficiency Evaluation}
To answer \textbf{RQ2}, we instigate into the relationship between the exploration quantity and the average attack success rates during the training process of DRL agents. As shown in Fig.~\ref{curve}, each unit of exploration quantity on the x-axis is $400 \times \frac{S}{\Delta_f}$, with $S$ representing the target steps of experience collection in each epoch. 

Furthermore, The minimal exploration quantity that G$^2$-SNIA requires to surpass the best baseline is also pointed out in the figure. For instance, as shown in Fig~\subref*{citeseer}, the red curve that represents the attack on the GCN model by G$^2$-SNIA on the Citeseer dataset exceeds the best baseline (the point where the dotted lines intersect in the figure) after 4 evaluation epochs, where we set $S$ to 4096, $\Delta_f$ to 54, $|V_{tar}|$ to 1000, and $|\mathcal{Y}|$ to 6. It can be estimated that the average exploration quantity for attacking each node and each label is only $\frac{4 \times 400}{1000 \times 6} \times \frac{4096}{54} \approx 21$, which is minimal compared to the entire search space of $\binom{F}{\Delta_f}=\binom{3327}{54}$. Additionally, we found that among these three datasets, GCN consistently requires the most exploration quantity to surpass the baseline method compared to other models. 

This observation implies that despite the baselines disrupting the model classification results through transfer attack, it is still capable of achieving satisfactory outcomes on the GCN model. In contrast, the other three models manage to match the performance of the baseline method with a smaller search effort.

\subsection{Budget Analysis}
To answer \textbf{RQ3}, we document the average probability change of target nodes being classified into targeted labels by victim models under varying attack budgets, as depicted in Fig.~\ref{line_graph}. The x-axis represents the different attack budgets, while the y-axis denotes the average probability, and the red dotted line signifies the attack budget established in our previous experiments. We observe that the curves in the graph are almost monotonically increasing, implying that as the attack budget grows, the likelihood of target nodes being classified as targeted labels by victim models increases. According to the changes in the slope of the curve, we find that as the attack budget increases, the slope of the curve tends to decrease, indicating a gradual reduction in the average probability change. This phenomenon indicates that our method initially selects features with a greater impact, allowing the confidence of targeted labels to rise rapidly during the early stages of the attack process. Additionally, even though we set fixed attack budgets while training the DRL agents, the results in the graph show that all curves maintain monotonicity even when exceeding one times the attack budget without requiring retraining, which implies that DRL agents can still ensure the quality of their decisions. In most datasets and models, the curves of different targeted labels are relatively concentrated. However, the various curves exhibit distinct growth trends in Fig.~\subref*{cora_gcnii}, \subref*{citeseer_gcnii} and \subref*{dblp_tagcn}, warranting further investigation into the cause of these differences in our future work.
\section{Conclusion} \label{conclusion}

In this work, we investigate a gradient-free single-node injection label specificity attack for graphs in the black-box evasion setting. We propose G$^2$-SNIA, a gradient-free, generalizable adversary that eliminates the risk of error propagation due to inaccurate approximations of the victim model. In contrast to other node injectors that require gradients from the surrogate model, G$^2$-SNIA operates without any assumptions about victim models. We formulate the single node injection label specificity attack as an MDP and solve it using a reinforcement learning framework. Through extensive experiments on three widely recognized datasets and four diverse GNNs, we demonstrate the promising performance of G$^2$-SNIA in comparison to state-of-the-art baselines in the black-box setting. To further showcase the effectiveness of G$^2$-SNIA, we also compare its attack performance with baselines in the white-box evasion setting, demonstrating that even under black-box conditions, our proposed method can still find solutions that are on par with the baselines.

Although G$^2$-SNIA demonstrates effectiveness and efficiency, there remain several challenges to address in future work. From the perspective of adversaries, as multi-agent reinforcement learning continues to advance, it is crucial to investigate how cooperation and competition between multiple injection nodes can be employed to implement label specificity attack on numerous nodes, ultimately yielding more sophisticated and covert offensive measures. Conversely, from a defensive standpoint, it is imperative to ascertain the extent of G$^2$-SNIA's effectiveness when applied to the robust GNNs, as well as to identify the salient architectural components of GNNs that can effectively withstand both GMA and GIA onslaughts. We will explore these issues in future research.

\section*{Acknowledgments}
This work was supported in part by the Key R\&D Program of Zhejiang under Grant 2022C01018 and 2021C01117, by the National Natural Science Foundation of China under Grants 61973273, 62103374 and U21B2001, by the National Key R\&D Program of China under Grant 2020YFB1006104, and by The Major Key Project of PCL under Grants PCL2022A03, PCL2021A02, and PCL2021A09.
\bibliographystyle{IEEEtran}
\bibliography{reference}
% \begin{IEEEbiography}[{\includegraphics[width=1in,height=1.25in,clip,keepaspectratio]{fig1.png}}]{IEEE Publications Technology Team}
% In this paragraph you can place your educational, professional background and research and other interests.\end{IEEEbiography}
  % \vspace*{-2.5cm}
\begin{IEEEbiography}[{\includegraphics[width=1in,height=1.25in,clip,keepaspectratio]{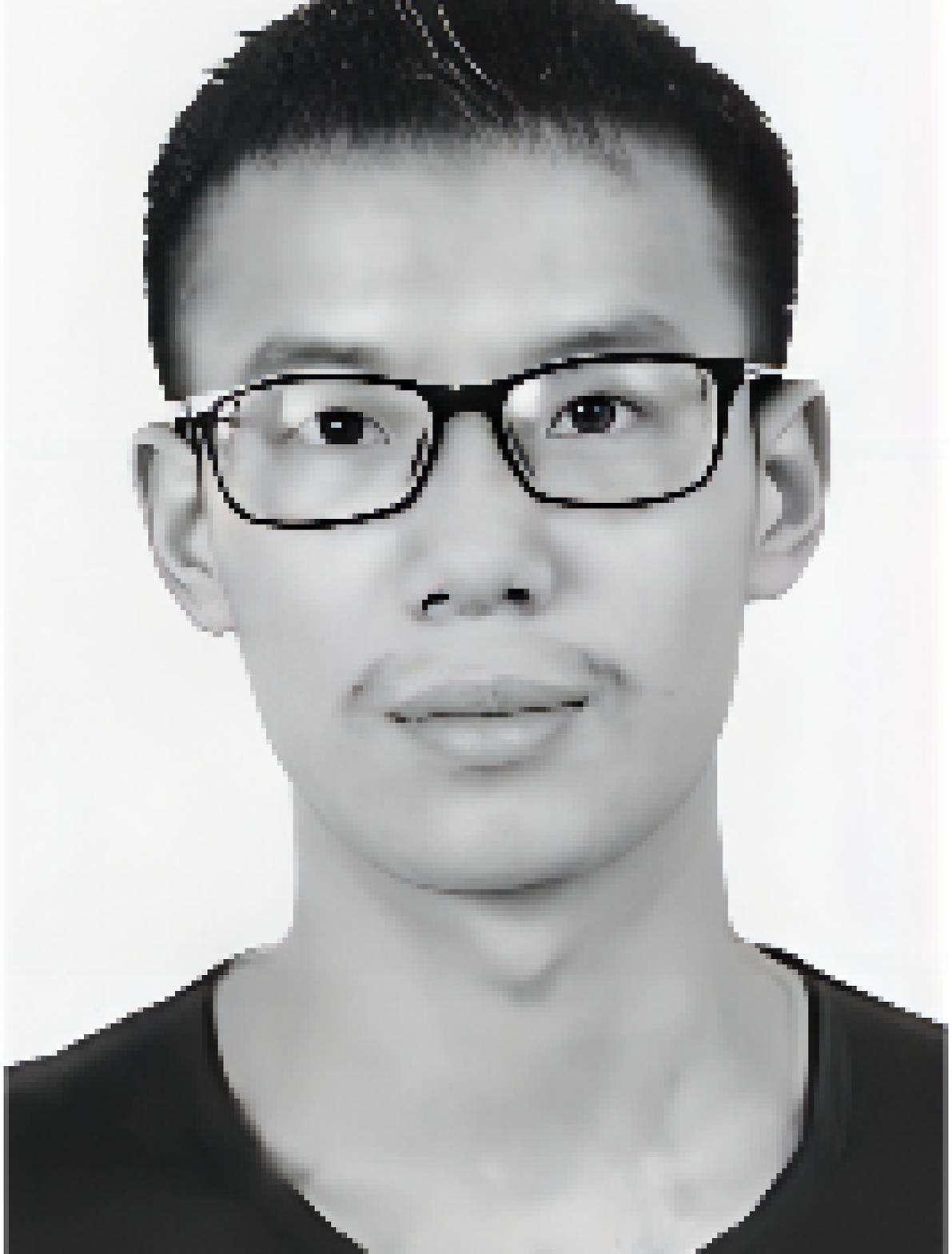}}]{Dayuan Chen}~received the BS degree in the College of Information Engineering, Zhejiang University of Technology, Hangzhou, China, in 2021. He is currently working toward the MS degree in the College of Information Engineering, Zhejiang University of Technology. His research is about graph machine learning security and reinforcement learning.
\end{IEEEbiography}

\vspace*{-1cm}
\begin{IEEEbiography}[{\includegraphics[width=1in,height=1.25in,clip,keepaspectratio]{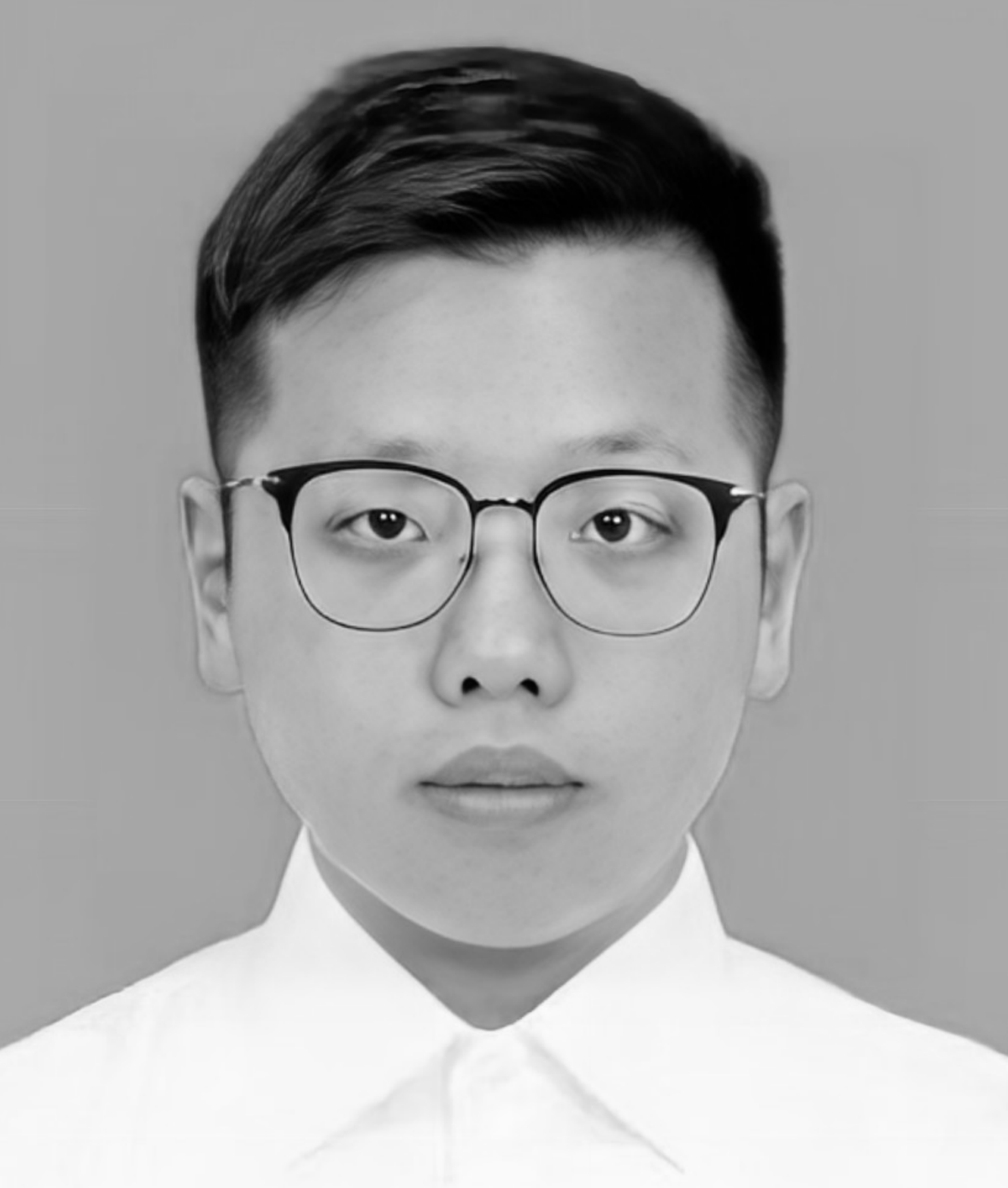}}]{Jian Zhang}~Jian Zhang received BS degree in automation and PhD degree in control theory and engineering from Zhejiang University of Technology, Hangzhou, China, in 2017 and 2022, respectively. He now is a Lecturer in the School of Cyberspace, China. His research interests include graph neural networks, machine learning security and anomaly detection.
\end{IEEEbiography}

\vspace*{-1cm}
\begin{IEEEbiography}[{\includegraphics[width=1in,height=1.25in,clip,keepaspectratio]{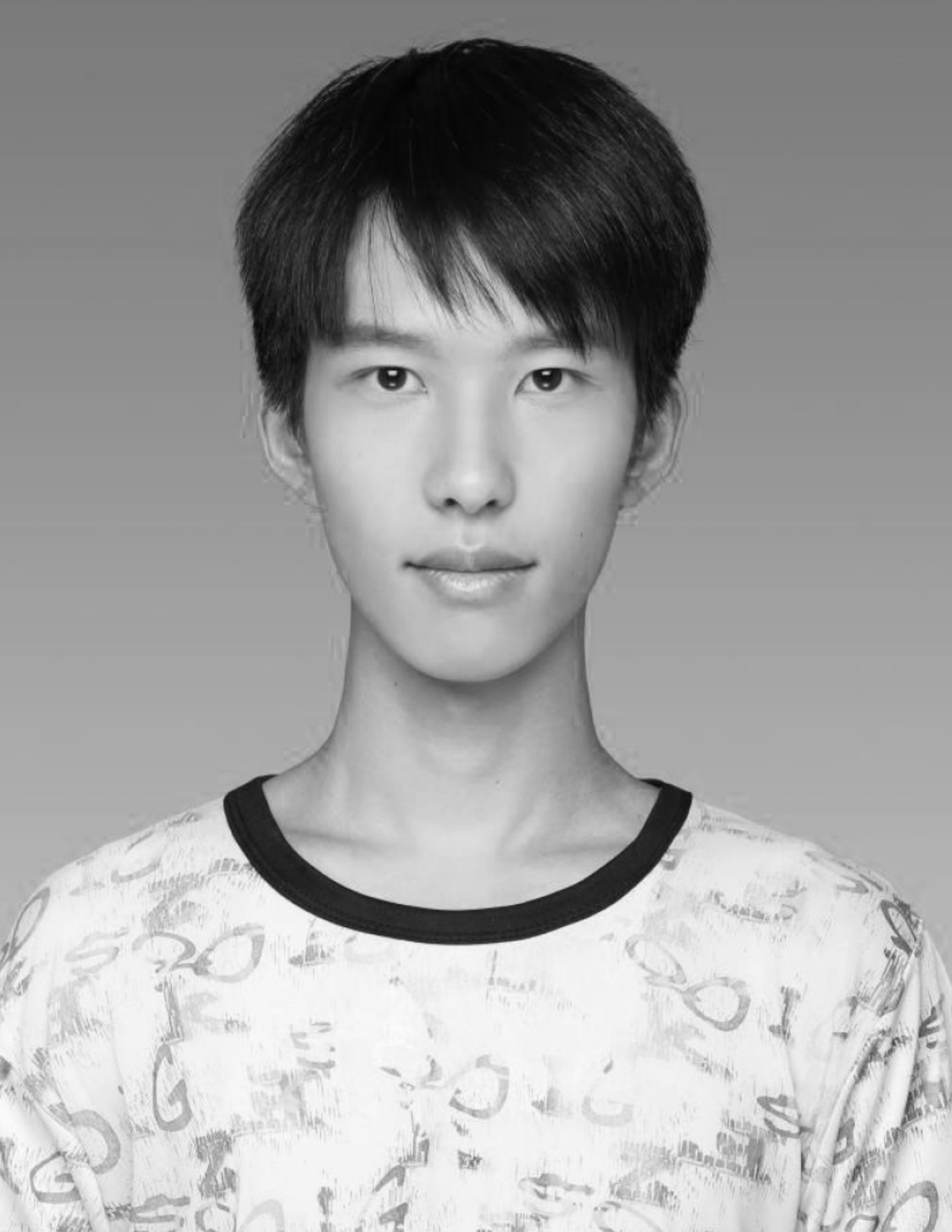}}]{Yuqian Lv}~received the BS degree in the College of Information Engineering, Zhejiang University of Technology, Hangzhou, China, in 2021. He is currently working toward the MS degree in the College of Information Engineering, Zhejiang University of Technology. His research is about node importance of social networks and machine learning.
\end{IEEEbiography}

\vspace*{-1cm}
\begin{IEEEbiography}[{\includegraphics[width=1in,height=1.25in,clip,keepaspectratio]{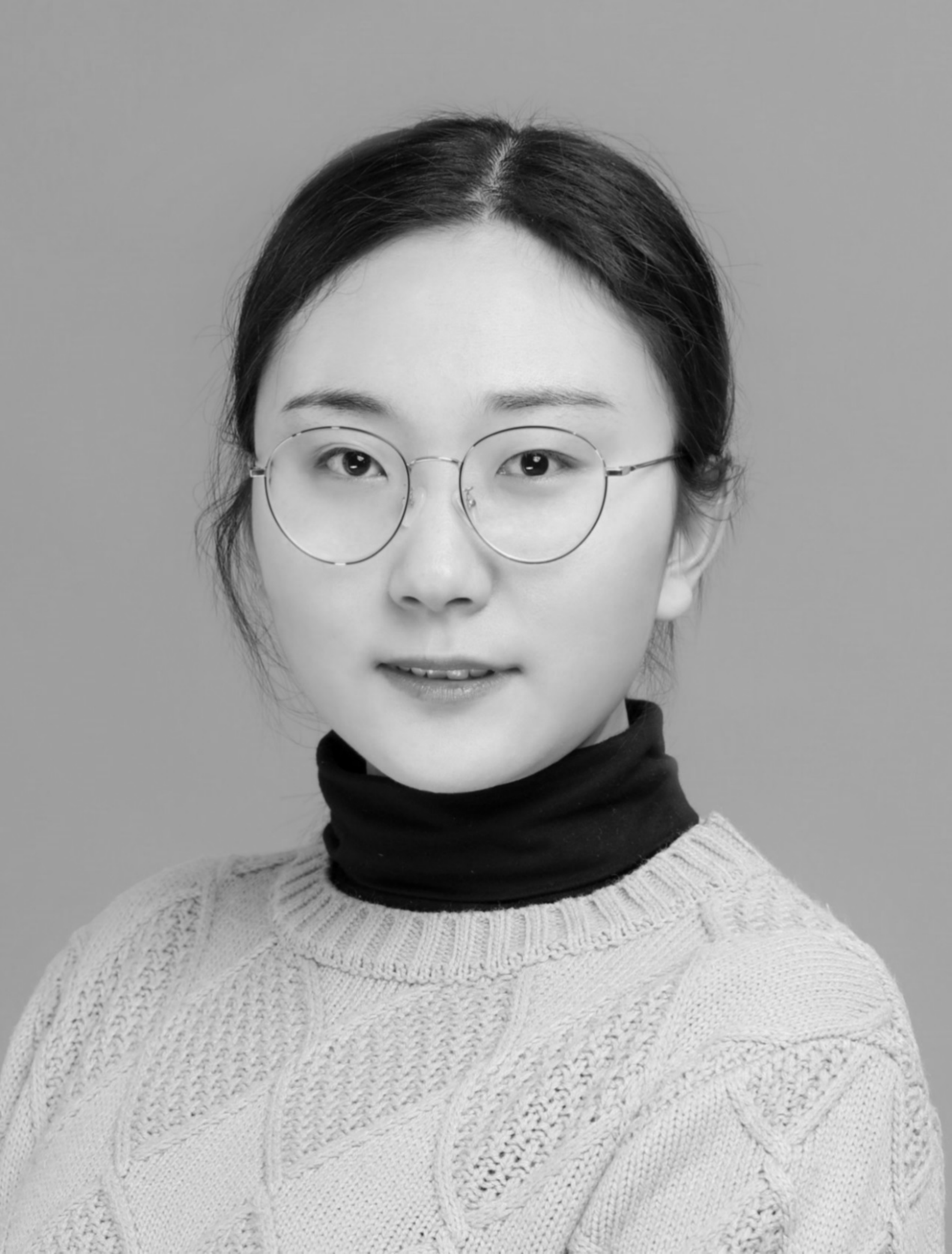}}]{Jinhuan Wang} received the BS and MS degree in the College of Information Engineering, Zhejiang University of Technology, Hangzhou, China, in 2017 and 2020. She is working toward the PhD degree in the College of Information Engineering, Zhejiang University of Technology. Her research interests include social network data mining and machine learning.
\end{IEEEbiography}

\vspace*{-1cm}
\begin{IEEEbiography}[{\includegraphics[width=1in,height=1.25in,clip,keepaspectratio]{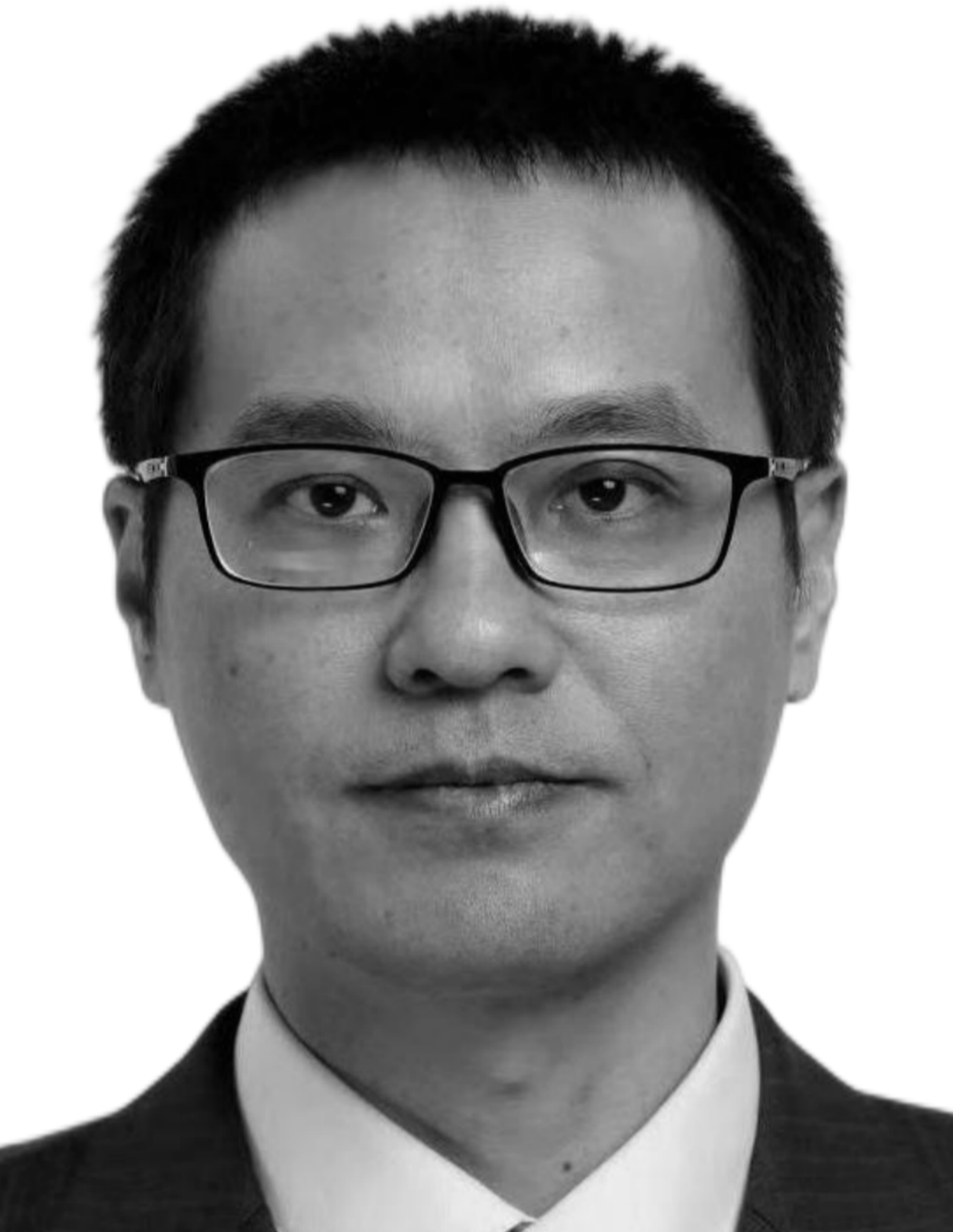}}]{Hongjie Ni}~received the B.S. degree in mechatronics from Hangzhou university of electronic science and technology, China, in 2003, and the M.S. degrees in Control Engineering from Zhejiang University of Technology, China, in 2008. He is currently a Senior engineer in College of Information Engineering, Zhejiang University of Technology. He has been focused on the integration of culture, science and technology since 2012, an acceptance expert for Ministry of science and technology "12th five-year" national science and technology support plan project, China. His current research interests include Mechatronics, intelligent control of stage equipment.
\end{IEEEbiography}

\vspace*{-1.5cm}
\begin{IEEEbiography}[{\includegraphics[width=1in,height=1.25in,clip,keepaspectratio]{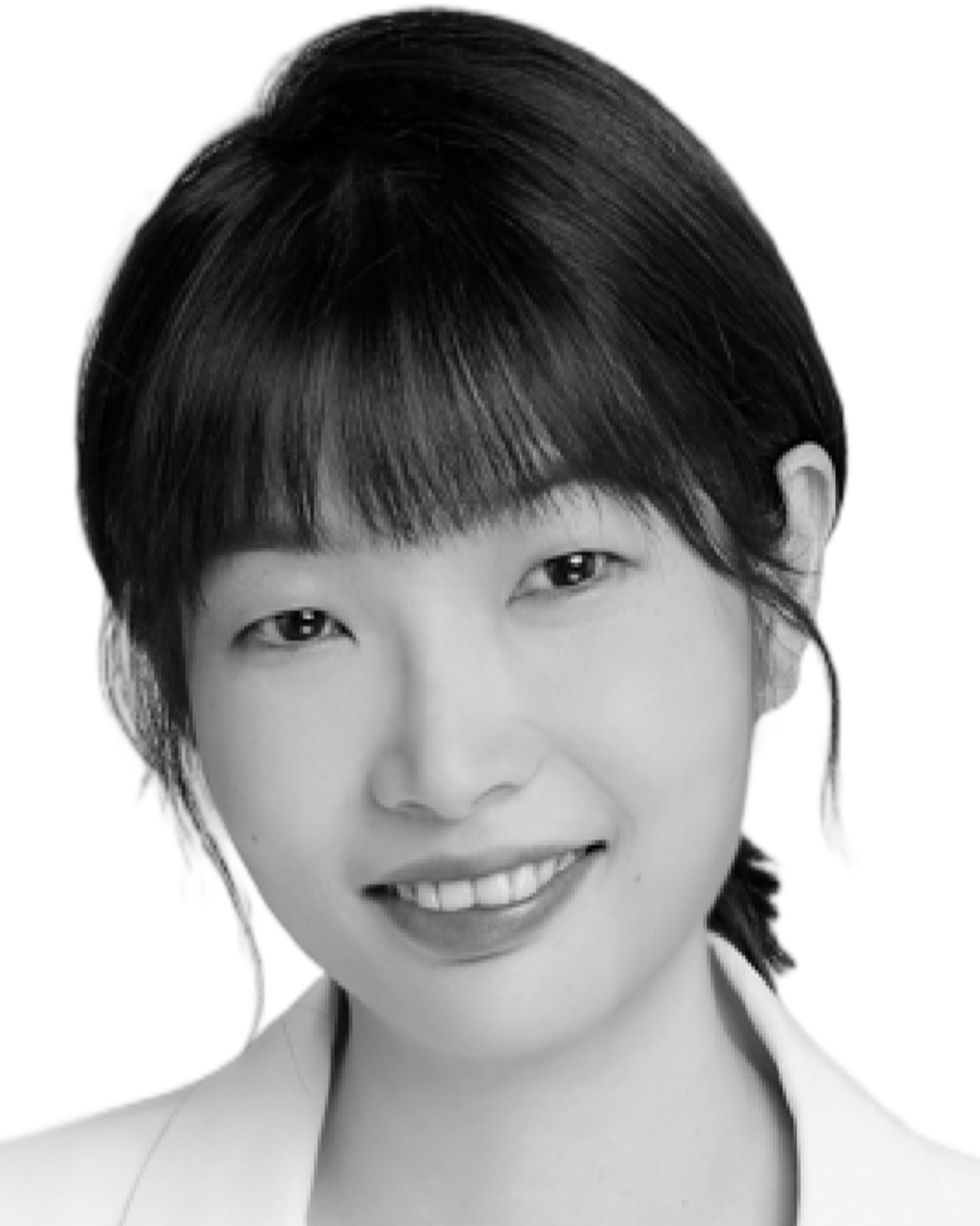}}]{Shanqing Yu}~received the M.S. degree from the School of Computer Engineering and Science, Shanghai University, China, in 2008 and received the M.S. degree from the Graduate School of Information, Production and Systems, Waseda University, Japan, in 2008, and the Ph.D. degree, in 2011, respectively. She is currently a Lecturer at the Institute of Cyberspace Security and the College of Information Engineering, Zhejiang University of Technology, Hangzhou, China. Her research interests cover intelligent computation and data mining.
\end{IEEEbiography}

\vspace{-2.5cm}
\begin{IEEEbiography}[{\includegraphics[width=1in,height=1.25in,clip,keepaspectratio]{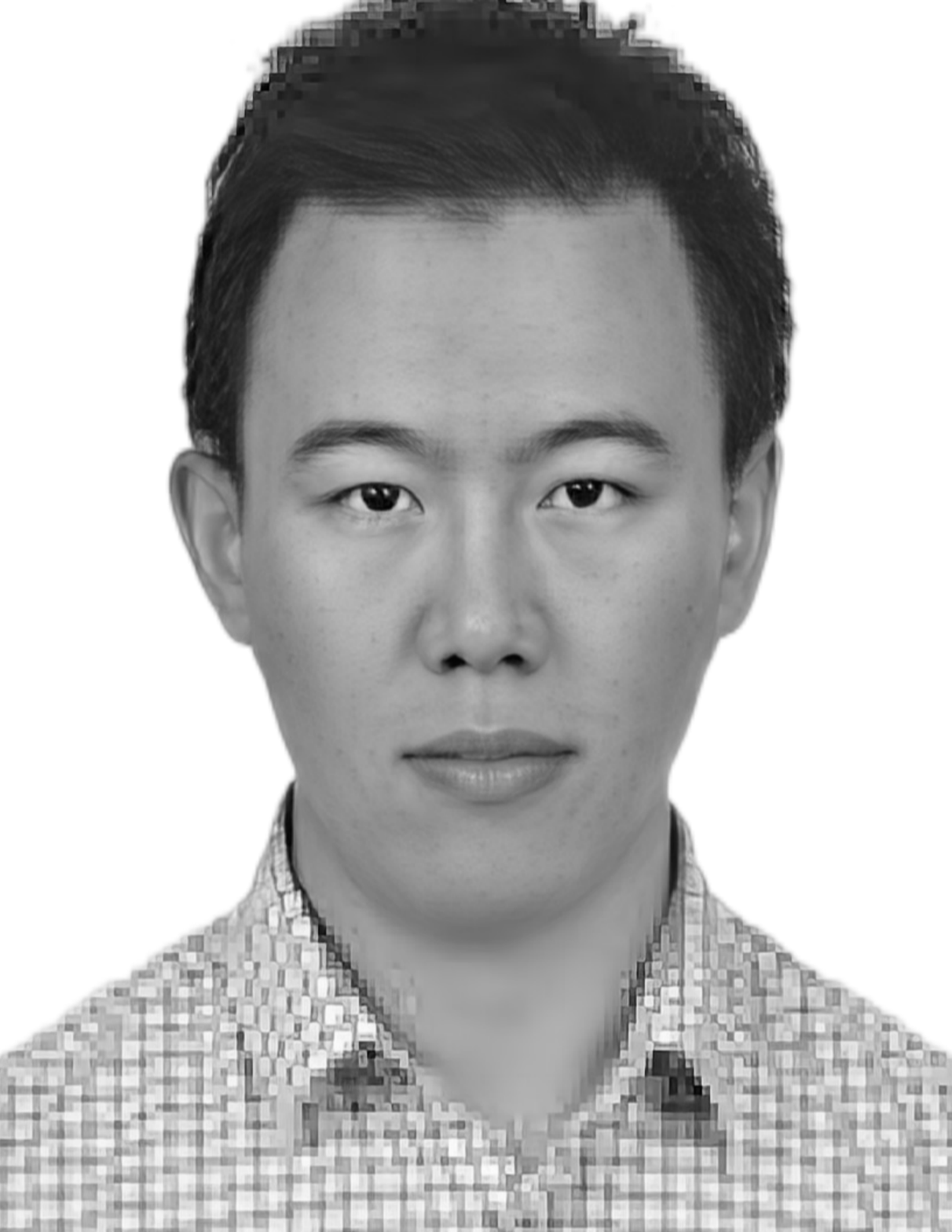}}]{Zhen Wang}~Zhen Wang is professor with School of Cyberspace, and vice dean of ZhuoYue Honors College, Hangzhou Dianzi University, China. He received BSc, MEng, PhD degree in Software Engineering from Dalian University of Technology, China, in 2007, 2009, and 2016. From 2014 to 2016, he was a research fellow at Nanyang Technological University, Singapore. His current research interests include: network security, artificial intelligence security, complex networks, reinforcement learning and algorithmic game theory.
\end{IEEEbiography}

\vspace{-3cm}
\begin{IEEEbiography}[{\includegraphics[width=1in,height=1.25in,clip,keepaspectratio]{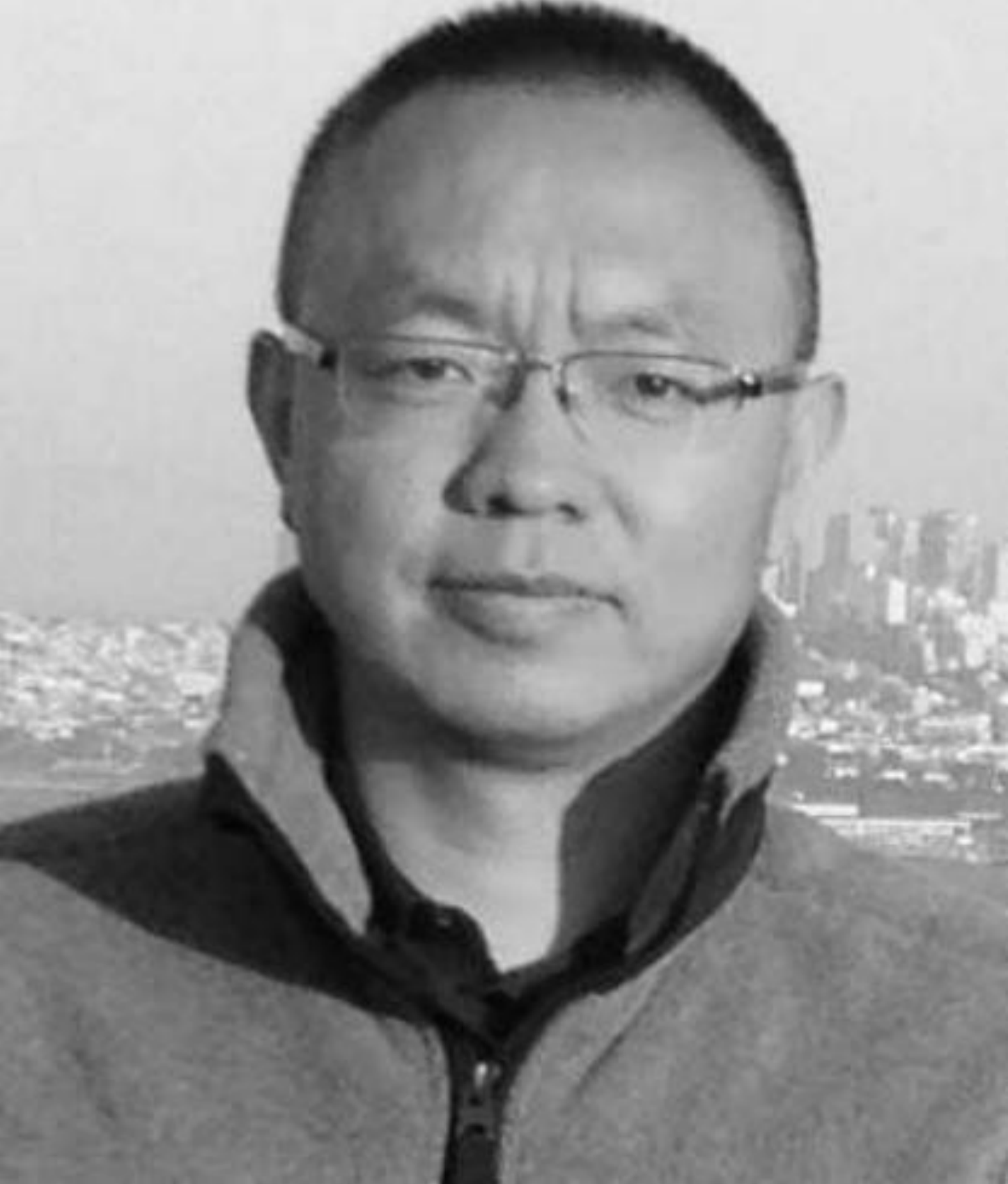}}]{Qi Xuan}~(M'18)~received the BS and PhD degrees in control theory and engineering from Zhejiang University, Hangzhou, China, in 2003 and 2008, respectively. He was a Post-Doctoral Researcher with the Department of Information Science and Electronic Engineering, Zhejiang University, from 2008 to 2010, and a Research Assistant with the Department of Electronic Engineering, City University of Hong Kong, Hong Kong, in 2010 and 2017. From 2012 to 2014, he was a Post-Doctoral Fellow with the
Department of Computer Science, University of California at Davis, CA, USA. He is a senior member of the IEEE and is currently a Professor with the Institute of Cyberspace Security, College of Information Engineering, Zhejiang University of Technology, Hangzhou, China. His current research interests include network science, graph data mining, cyberspace security, machine learning, and computer vision. 
\end{IEEEbiography}

\vspace{8.5cm}

\end{document}